\documentclass[11pt]{article}

\usepackage[final]{acl}

\usepackage{times}
\usepackage{latexsym}
\usepackage{amsthm}
\usepackage[T1]{fontenc}

\usepackage[utf8]{inputenc}

\usepackage{microtype}

\usepackage{inconsolata}

\usepackage{amsmath, amssymb, amsfonts} 
\usepackage{graphicx} 
\usepackage{multirow} 
\usepackage{booktabs} 
\usepackage{bm} 
\usepackage{mathtools} 
\usepackage{lipsum} 
\usepackage{ragged2e} 
\usepackage{subcaption} 
\usepackage{threeparttable} 
\usepackage{tabularx}
\usepackage{array}

\usepackage{geometry}
\usepackage[table]{xcolor}

\definecolor{header_blue}{HTML}{E3F2FD} 

\title{Modeling Multi-Dimensional Cognitive States in Large Language Models under Cognitive Crowding}

\author{Lin Zhong$^{1}$\thanks{~Equal contribution.}, Siyu Zhu$^{1}$\footnotemark[1], Zizhen Yuan$^{1}$, Jinhao Cui$^{1}$, Xinyang Zhao$^{1}$, \\
  \textbf{Lingzhi Wang}$^{1}$, \textbf{Hao Chen}$^{2}$, \textbf{Qing Liao}$^{1,3}$\thanks{~Corresponding author.} \\
  $^{1}$Harbin Institute of Technology, Shenzhen, China \quad
  $^{2}$City University of Macau, Macao SAR, China \\
  $^{3}$Peng Cheng Laboratory, Shenzhen, China \\
  {\small\texttt{\{zhonglin,siyuzhu,cuijinhao,25S151188,25S151205\}@stu.hit.edu.cn}} \\
  {\small\texttt{\{wanglingzhi,liaoqing\}@hit.edu.cn},\ \texttt{sundaychenhao@gmail.com}}}

\begin{document}
\maketitle

\begin{abstract}
Modeling human cognitive states is essential for advanced artificial intelligence. Existing Large Language Models (LLMs) mainly address isolated tasks such as emotion analysis or stance detection, and fail to capture interactions among cognitive dimensions defined in psychology, including emotion, thinking style, stance, and intention. To bridge this gap, we construct CognitiveBench, the first benchmark with unified annotations across the above four dimensions. Experiments on CognitiveBench show that although LLMs perform well on single dimension tasks, their performance drops sharply in joint multi-dimensional modeling. Using Gromov $\delta$-hyperbolicity analysis, we find that CognitiveBench exhibits a strong hierarchical structure. We attribute the performance bottleneck to ``Cognitive Crowding'', where hierarchical cognitive states require exponential representational space, while the Euclidean space of LLMs grows only polynomially, causing representation overlap and degraded performance. To address this mismatch, we propose \textbf{HyCoLLM}, which models cognitive states in hyperbolic space and aligns LLM representations via Hyperbolic Guided Alignment Tuning. Results show that HyCoLLM substantially improves multi-dimensional cognitive understanding, allowing 8B parameter model to outperform strong baselines, including GPT-4o. Our code, dataset, and trained weights are available at \url{https://github.com/Chips98/HyCoLLM_for_ACL2026}.
\end{abstract}

\section{Introduction}

\begin{figure}[t]
    \centering
    \includegraphics[width=0.90\linewidth]{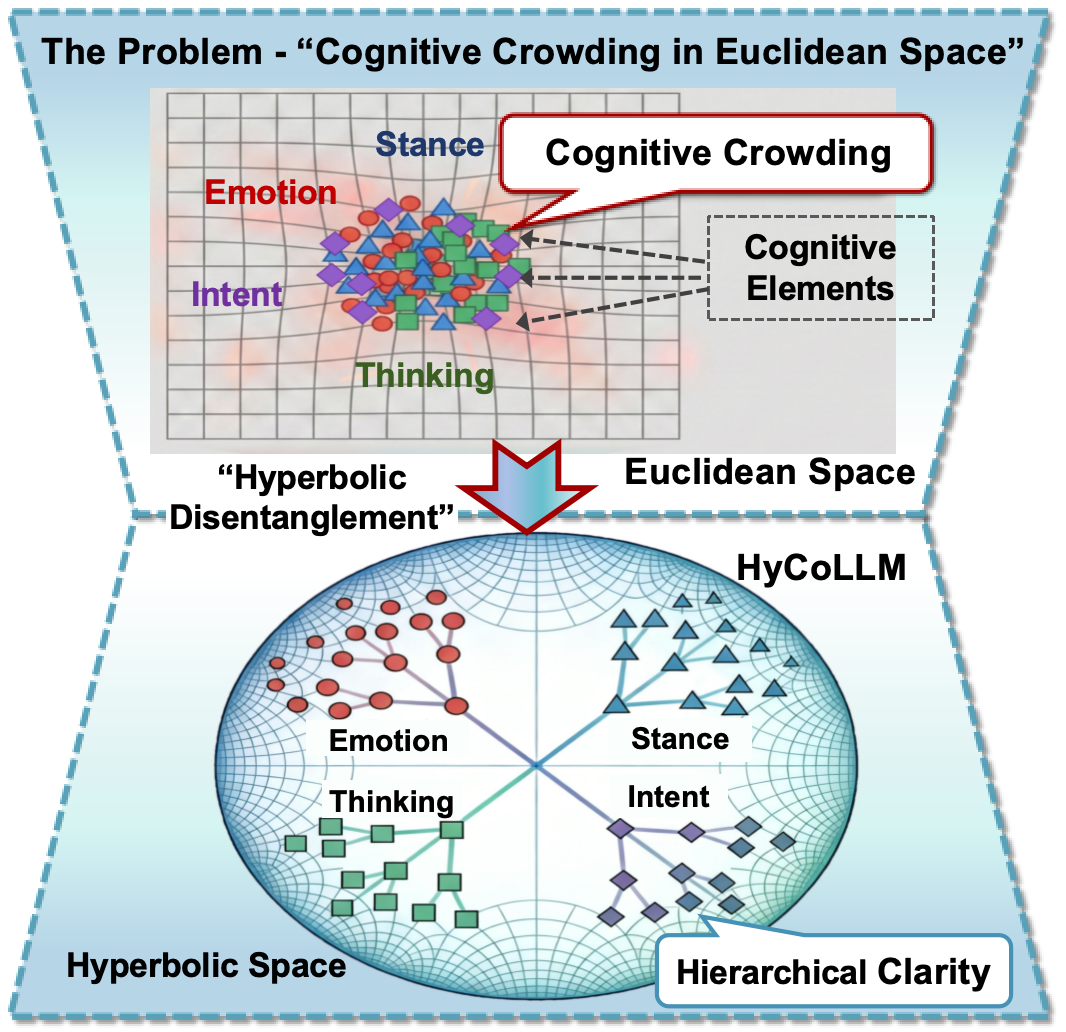} 
    \caption{Illustration of Alleviating Cognitive Crowding in Euclidean Space through Hyperbolic Disentanglement.}
    \label{fig_intro_1}
\end{figure}

Understanding human cognitive states is a fundamental goal of artificial intelligence, with applications in psychological counseling, social media analysis, and dialogue systemsc \citep{wang2025llm4deu, zhong2025comapoi}. Although Large Language Models (LLMs) have shown promising capabilities in this area, most existing approaches treat psychological dimensions in isolation. Prior work typically focuses on single tasks such as stance detection \citep{li2021p}, emotion analysis \citep{mohammad2016semeval}, or intent recognition \citep{mirzaei2023real}. This task specific perspective oversimplifies human cognition. Psychological research suggests that cognitive dimensions form an integrated system with interactions \citep{eagly1993psychology,breckler1984empirical}. For instance, an opposing stance may arise from a deliberative thinking style or from emotions such as anger. Ignoring these interactions prevents LLMs from capturing a complete cognitive state, which jointly involves emotion, thinking style (abbreviated as thinking), stance, and intent \citep{shapira2024clever}. This gap highlights the need for benchmarks that support holistic cognitive modeling across multiple dimensions.

Existing benchmarks fall short of this requirement. Some datasets annotate two dimensions jointly, such as stance and emotion \citep{pu2025integrating, upadhyaya2023multi, gambini2024evaluating}, or incorporate intent information in dialogue settings \citep{donmez2025understand, mirzaei2023real}. However, they rarely cover the above four dimensions simultaneously. More importantly, most benchmarks lack explicit annotations for thinking, which plays a central role in linking emotion to stance. To address this limitation, we construct CognitiveBench, the first dataset with fine grained annotations for emotion, thinking, stance, and intent. CognitiveBench enables a systematic evaluation of whether LLMs can capture the holistic cognitive state rather than isolated components.

Using CognitiveBench, we first evaluate the performance of current LLMs on multi-dimensional cognitive understanding. The results reveal a clear limitation: while LLMs achieve strong performance on individual dimensions, their accuracy drops sharply when required to jointly predict all four dimensions. For example, GPT-4o performs reasonably well in a single dimension, but when it models four cognitive dimensions simultaneously, its accuracy drops to approximately 5.7\%. This failure motivates a deeper analysis of the underlying data structure. We analyze CognitiveBench using Gromov $\delta$-hyperbolicity, which measures the extent to which data exhibit tree like structure. The analysis shows that CognitiveBench has a pronounced intrinsic hierarchy, with relative $\delta \approx 1\%$. 

Psychological studies also indicate that cognitive states exhibit a pronounced hierarchical structure \citep{shaver1987emotion, taylor2015global}, which is consistent with our findings from the Gromov $\delta$-hyperbolicity analysis. We therefore attribute the performance bottleneck of LLMs to this finding, which we term "Cognitive Crowding". It refers to the mismatch between the representational demands of cognitive states and the geometry of LLM embeddings. Hierarchical cognitive states require a space whose volume grows exponentially, whereas standard LLMs rely on Euclidean representation spaces with only polynomial growth. As a result, distinct cognitive states collapse and overlap in Euclidean space, preventing LLMs from effectively distinguishing them.

To address this geometric mismatch, we propose the HyCoLLM framework. HyCoLLM models cognitive states in hyperbolic space, specifically using the Poincaré ball model, which naturally supports exponential volume growth and hierarchical structure. We first design a Hyperbolic Cognitive Network (HCN) that separates cognitive states through a geometry aware contrastive loss. We then propose Hyperbolic Guided Alignment Tuning (HGAT), which aligns the internal representations of an LLM with the learned hyperbolic cognitive manifold via semantic-cognitive topology loss. The main contributions are as follows:

\begin{itemize}
    \setlength{\itemsep}{0pt}
    \setlength{\parsep}{0pt}
    \setlength{\parskip}{0pt}
    \item We introduce CognitiveBench, the first four-dimensional cognitive benchmark covering emotion, thinking, stance, and intent, enabling holistic evaluation of cognitive understanding.
    \item We identify Cognitive Crowding as a key obstacle in multi-dimensional cognitive modeling with LLMs, and provide empirical evidence through performance analysis and topological characterization of the data.
    \item We propose the HyCoLLM framework, which integrates hyperbolic geometric priors with an alignment based tuning strategy to mitigate cognitive crowding.
    \item Experimental results show that HyCoLLM substantially improves cognitive understanding across both single and joint tasks, with an 8B parameter model outperforming strong baselines, including GPT-4o.
\end{itemize}

\section{Related Work}

\paragraph{Multi-dimensional Cognitive Benchmarks.}
The landscape of cognitive dataset construction has evolved from single-task annotations, such as P-Stance~\citep{li2021p}, MultiTarget~\citep{sobhani2017dataset}, and ASTSD~\citep{wan2015ensemble}, toward multi-dimensional modeling. Recognizing the inherent correlations between mental states, some benchmarks have begun to bridge distinct dimensions. For instance, SemEval-16~\citep{mohammad2016semeval} established a benchmark for the joint detection of stance and sentiment, while GunStance~\citep{gyawali2024gunstance} and EcoVerse~\citep{grassoecoverse} extended this synergy to controversial social topics. Similarly, in intent modeling, QID~\citep{mirzaei2023real} captured the intent behind questions , and datasets derived from ChangeMyView~\citep{donmez2025understand} investigated persuasive intent. However, as presented in Table \ref{tab:dataset_comparison}, although these datasets have attempted to cover two cognitive dimensions, they still cannot fully reflect the overall cognitive state of the speaker. For this reason, CognitiveBench has supplemented more dimensions.

\paragraph{Modeling Cognitive States in LLMs.}
Prior research has leveraged LLMs to model specific psychological dimensions, including emotion, stance, and intent~\citep{amirizaniani2024can}. In \textit{Emotion Analysis}, approaches have advanced from simple categorization to context-aware and multimodal modeling~\citep{kluwer2025context, fazzi2025don, yu2024towards}. For \textit{Stance Detection}, methods largely mitigate data scarcity through background knowledge extraction~\citep{zhang2024llm} and chain-of-thought augmentation~\citep{zou2025tat}. Similarly, \textit{intent Modeling} has focused on user-centric benchmarks~\citep{wang2024user} and applications in cognitive behavioral therapy~\citep{na2024cbt, xu2025integrating}. Despite these advancements, these works typically model each dimension in isolation within Euclidean spaces~\citep{yadav2024pag, akash2025can}. Such approaches overlook the intrinsic hierarchical synergy of complex mental states, failing to capture the structural dependencies that organize human cognition.


\begin{table}[t]
\centering
\resizebox{0.49\textwidth}{!}{%
\begin{tabular}{l c l cccc}
\toprule
\textbf{Dataset} & \textbf{Size} & \textbf{Domain} & \textbf{Sta.} & \textbf{Emo.} & \textbf{Int.} & \textbf{Thk.} \\
\midrule
Covid19-Stance & 7,122 & Covid-19 & \checkmark & \checkmark & $\times$ & $\times$ \\
MultiTarget & 4,455 & Political & \checkmark & $\times$ & $\times$ & $\times$ \\
SemEval-16& 4,163 & Political & \checkmark & \checkmark & $\times$ & $\times$ \\
GunStance & 5,500 & Gun Policy & \checkmark & $\times$ & $\times$ & $\times$ \\
QID & 2,084 & Social Life & $\times$ & $\times$ & \checkmark & $\times$ \\
ASTSD & 12,864 & Airline & $\times$ & \checkmark & $\times$ & $\times$ \\
\rowcolor{gray!10} \textbf{CognitiveBench} & \textbf{6,514} & \textbf{Pol./Eco./Cul.} & \textbf{\checkmark} & \textbf{\checkmark} & \textbf{\checkmark} & \textbf{\checkmark} \\
\bottomrule
\end{tabular}%
}
\caption{Comparison of CognitiveBench with existing representative datasets. While existing datasets typically focus on one or two specific dimensions, CognitiveBench integrates four synergistic dimensions: Stance, Emotion, Intent, and Thinking.}
\label{tab:dataset_comparison}
\end{table}

\section{Dataset: CognitiveBench}
\label{sec:cognitive_bench}

To investigate the interaction between different cognitive dimensions, we need a benchmark that captures the full complexity of human cognition. As shown in Table~\ref{tab:dataset_comparison}, existing datasets typically cover only one or two dimensions, which restricts the study of how these states influence each other. To address this, we construct \textbf{CognitiveBench}, the first dataset to jointly model emotion, thinking, stance, and intent. In this work, we define a ``cognitive state'' as the specific combination of these four dimensions.

The taxonomy of each dimension is grounded in established cognitive theories. Emotions are organized following Plutchik’s model~\citep{plutchik1980emotion}, with positive, negative, and neutral categories and 9 basic emotions. Thinking is divided into intuitive and analytical types inspired by dual process theory~\citep{kahneman2011thinking}, and further specified into 8 concrete thinking styles. User stance towards a target is defined based on social judgment theory~\citep{sherif1961social} and adapted to each topic, since stance targets vary across domains. User intent is categorized into representatives, directives, and expressives based on speech act theory~\citep{austin1975things}, yielding 7 fine grained intent labels. The overview of labels is provided in Table~\ref{tab:taxonomy_overview}, with details in Appendix~\ref{app:definitions}.

\begin{table}[t]
\centering
\Large
\resizebox{0.48\textwidth}{!}{%
\begin{tabular}{l l m{7cm}}
\toprule
\textbf{Dimension} & \textbf{Category} & \textbf{Labels} \\ \midrule
\multirow{3}{*}{Emotion} 
 & Positive & Joy, Trust, Anticipation, Surprise \\
 & Negative & Anger, Disgust, Fear, Sadness \\
 & Neutral  & Neutral \\ \midrule
\multirow{2}{*}{Thinking} 
 & Intuitive & Subjective Evaluation, Identity Conformity, Emotional Judgment, Experience-Based \\
 & Analytical & Logical, Balanced Consideration, Evidence-Based, Critical \\ \midrule
Stance 
 & N/A & Support, Oppose, Unclear \\ \midrule
\multirow{3}{*}{Intent} 
 & Representatives & Information Sharing, Opinion Expression \\
 & Directives      & Information Seeking, Call to Action \\
 & Expressives     & Connection, Conflict, Emotional Expression \\ \bottomrule
\end{tabular}%
}
\caption{Taxonomy of the four cognitive dimensions in CognitiveBench.}
\label{tab:taxonomy_overview}
\end{table}

\subsection{Data Construction}
\label{sec:data_construction}

We collect data from the Twitter (X) platform centering on four distinct topics: \textit{China-United States Trade (CUT)}, \textit{United States Election (UE)}, \textit{Diversity, Equity, and Inclusion (DEI)}, and \textit{Federal Reserve Interest Rates (FRIR)}. We select these diverse topics to verify model performance across different domains: CUT and FRIR represent the economic domain, UE represents the political domain, and DEI represents the cultural domain. Starting from a raw corpus of more than 48,000 posts, we conduct systematic data filtering and cleaning to ensure sample validity and research reliability. We remove short texts with insufficient semantic content, defined as fewer than 10 tokens. All user information is anonymized to protect privacy. We then manually review each remaining sample to assess topic relevance and remove off-topic content. After filtering, we obtain approximately 9,000 valid candidate samples.

\paragraph{Expert Annotation.}
To ensure high annotation quality, we recruit 29 domain experts with backgrounds in affective computing or psychology. The entire annotation campaign lasts for two months, with an hourly wage of \$45. The annotation process consists of four stages. First, five annotators conduct pilot annotation on 1,000 samples. Based on their annotations and feedback, we iteratively revise the annotation guidelines and label boundaries to establish clear and consistent definitions for all four dimensions. Second, all annotators receive training based on the revised guidelines. After training, each annotator labels a designated set of samples, and the research team evaluates their performance on 100 samples and conducts follow up discussions to ensure agreement on annotation standards. Third, each sample is independently annotated by three annotators across all four dimensions. Fourth, we retain only samples that reach majority agreement, defined as an agreement ratio of at least $2/3$, and discard samples with substantial disagreement.

\paragraph{Dataset Statistics and Reliability.}
Starting from a raw corpus of 48,496 posts, our multi-stage filtering and expert-consensus pipeline retains 6,514 high-quality samples (overall retention $\approx 13.4\%$), distributed across the four topics as 1,445 (CUT), 2,090 (UE), 1,227 (DEI), and 1,752 (FRIR). As summarized in Table~\ref{tab:dataset_stats}, the four domains differ in both raw volume and selectivity: the politically and economically charged UE and FRIR topics yield higher retention ($18.6\%$ and $18.0\%$), while the more contested CUT and DEI topics are filtered more aggressively ($9.8\%$ and $9.6\%$) to remove off-topic or low-signal utterances. The resulting corpus also spans a healthy breadth of contributors, with 283/309/82/317 unique users per topic and an average of 5-15 posts per user, providing diverse expressions of emotion, thinking style, stance, and intent rather than being dominated by a handful of prolific accounts. We split each topic into training and test sets following an 8:2 ratio, with label proportions preserved within each split to keep evaluation representative.

Annotation reliability is measured by Cohen's $\kappa$ computed on the three-way expert annotations for every dimension and every topic. All $\kappa$ values fall within the ``moderate'' to ``substantial'' range ($0.51$–$0.79$), indicating that even the finer-grained dimensions remain consistently identifiable by independent experts. The \emph{Stance} dimension attains the highest agreement (up to $\kappa=0.79$ on CUT), reflecting the relatively unambiguous support/oppose/unclear signals in political and economic discussions, whereas \emph{Emotion} and \emph{Intent} show slightly lower but still substantial agreement due to the inherently more subjective nature of affect attribution. We view this combination of scale, topical diversity, multi-user coverage, and verified annotation quality as sufficient to support a rigorous evaluation of multi-dimensional cognitive modeling. Additional breakdowns, including label-level distributions for each dimension, are provided in Appendix~\ref{app:annotation}.

\begin{table}[t]
\centering
\small
\setlength{\tabcolsep}{3pt}
\begin{tabular}{@{}l cccc@{}}
\toprule
\textbf{Topic} & \textbf{CUT} & \textbf{UE} & \textbf{DEI} & \textbf{FRIR} \\
\textit{(Domain)} & \textit{(Trade)} & \textit{(Politic)} & \textit{(Culture)} & \textit{(Economy)} \\ \midrule

\multicolumn{5}{@{}l}{\textbf{Data Statistics}} \\
\hspace{3mm} Raw Posts & 14,773 & 11,234 & 12,759 & 9,730 \\
\hspace{3mm} Valid Samples$^\dagger$ & 1,445 & 2,090 & 1,227 & 1,752 \\
\hspace{3mm} Retention & 9.8\% & 18.6\% & 9.6\% & 18.0\% \\ \midrule

\multicolumn{5}{@{}l}{\textbf{User Interaction}} \\
\hspace{3mm} Unique Users & 283 & 309 & 82 & 317 \\
\hspace{3mm} Posts/User & 5.1 & 6.8 & 15.0 & 5.5 \\ \midrule

\multicolumn{5}{@{}l}{\textbf{Reliability (Kappa $\kappa$)}} \\
\hspace{3mm} Emotion & 0.68 & 0.62 & 0.55 & 0.59 \\
\hspace{3mm} Thinking & 0.61 & 0.58 & 0.64 & 0.61 \\
\hspace{3mm} Stance & 0.79 & 0.71 & 0.51 & 0.63 \\
\hspace{3mm} Intent & 0.66 & 0.63 & 0.60 & 0.65 \\ \bottomrule
\end{tabular}

\vspace{1ex}
{\footnotesize \textit{Note:} $^\dagger$ Valid samples after filtering for length, toxicity, and consensus.}
\caption{Per-topic statistics and inter-annotator reliability of the \textit{CognitiveBench} dataset. We report raw volume, retention after filtering, user-level diversity, and Cohen's $\kappa$ for each cognitive dimension across the four topics.}
\label{tab:dataset_stats}
\end{table}

\subsection{Cognitive Crowding}

\begin{figure}[t]
    \centering
    \includegraphics[width=0.99\linewidth]{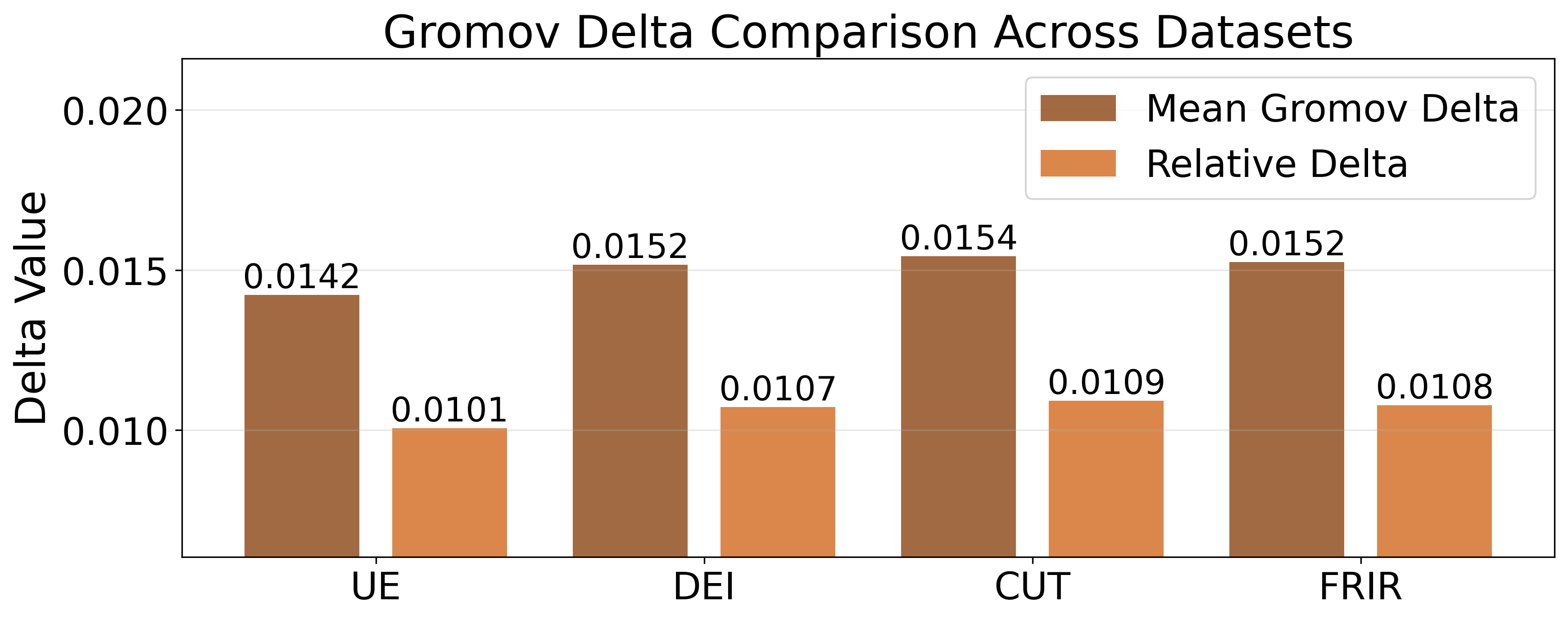} 
    \caption{Gromov $\delta$ hyperbolicity analysis across four datasets. Consistently low relative delta values of approximately 1\% indicate a strong intrinsic hierarchical structure.}
    \label{fig:gromov_analysis}
\end{figure}

\begin{figure*}[t]
    \centering
    \includegraphics[width=\linewidth]{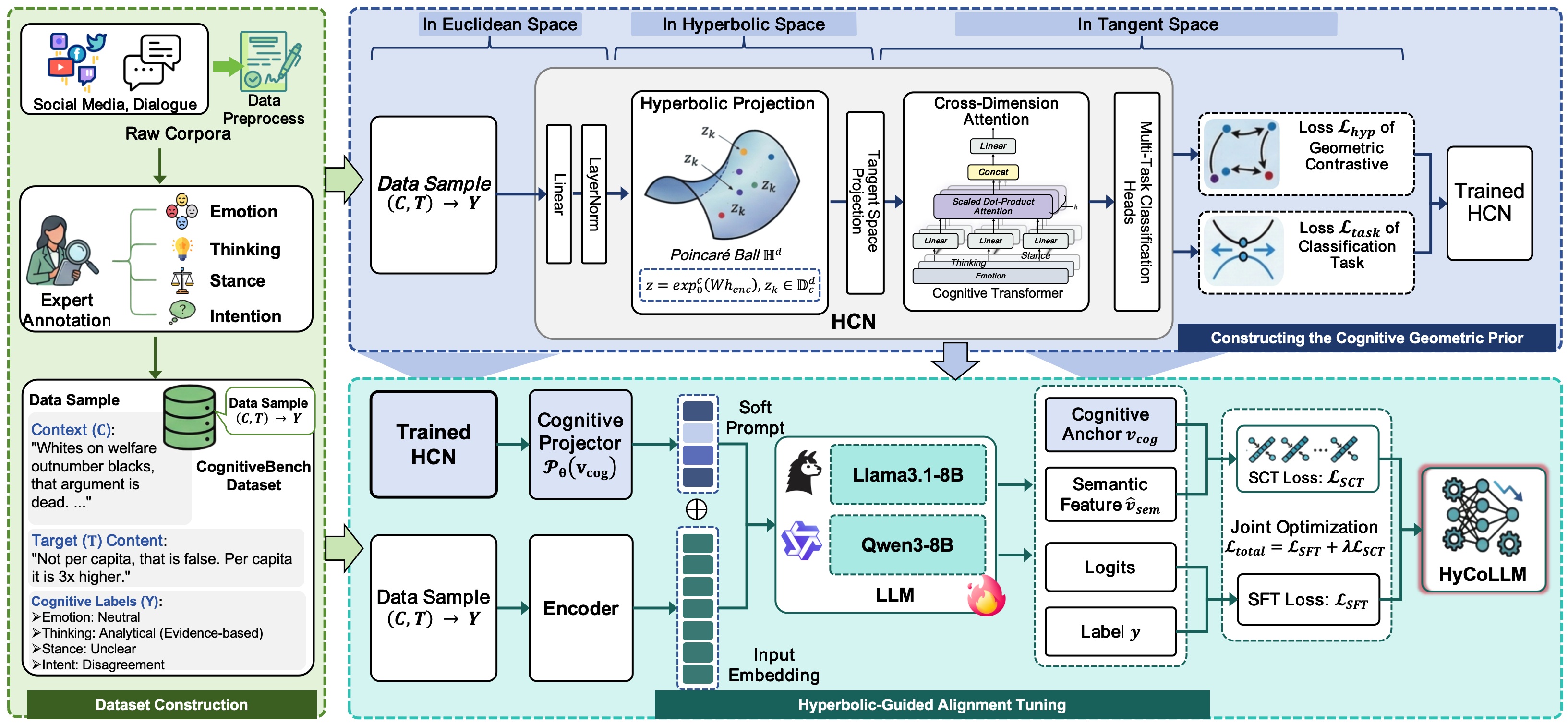} 
    \caption{The overall framework of HyCoLLM.}
    \label{fig:framework}
\end{figure*}
Based on CognitiveBench, we first investigate the capability of current LLMs to model complex cognitive states. We evaluate state-of-the-art models, including GPT-4o and Claude-4.5. While these models perform reasonably well on single dimensions, they fail significantly on the joint prediction task. As shown in Table~\ref{tab:main_results}, the ``all-correct'' accuracy (requiring correct predictions for emotion, thinking, stance, and intent simultaneously) is only 5.7\% for GPT-4o and 3.4\% for Claude-4.5. Analysis reveals that models frequently confuse concepts that are semantically related but hierarchically distinct, such as misidentifying ``intuitive thinking'' as ``emotional judgment.''

We hypothesize that this failure stems from ``Cognitive Crowding,'' a phenomenon caused by a geometric mismatch. Cognitive theories indicate that mental states are hierarchical, where broad categories branch into fine-grained concepts~\citep{shaver1987emotion, taylor2015global}. To confirm this structure in our data, we apply Gromov $\delta$-hyperbolicity analysis. As shown in Figure~\ref{fig:gromov_analysis}, the relative delta values across all domains are approximately 1\%, which indicates a strong intrinsic tree-like structure. Hierarchical structures require space that grows exponentially ($b^d$) to separate branches, whereas the Euclidean space used by LLMs only grows polynomially ($R^d$). Consequently, when LLMs attempt to embed this hierarchy into Euclidean space, distinct cognitive states are forced into overlapping regions. This geometric bottleneck creates the ``Cognitive Crowding'' effect, limiting the model's ability to distinguish multi-dimensional states.

\section{Method: HyCoLLM}
\label{sec:method}

To resolve the cognitive crowding problem observed in Section \ref{sec:cognitive_bench}, we propose \textbf{HyCoLLM} (Hyperbolic Cognitive Large Language Model). Instead of forcing hierarchical cognitive states into Euclidean space, HyCoLLM grounds the understanding capabilities of LLMs on a hyperbolic manifold. As illustrated in Figure \ref{fig:framework}, the framework operates in two phases: (1) disentangling multi-dimensional cognitive states using the Hyperbolic Cognitive Network (HCN), and (2) aligning LLM semantic representations with the cognitive geometric prior via Hyperbolic-Guided Alignment Tuning (HGAT).

\paragraph{Preliminaries: The Poincaré Ball.}
We employ the Poincaré ball model $(\mathbb{D}_c^d, g^{\mathbb{D}})$ to capture hierarchical structures. The manifold is defined as $\mathbb{D}_c^d = \{\boldsymbol{z} \in \mathbb{R}^d : \|\boldsymbol{z}\| < 1/\sqrt{c}\}$ with constant negative curvature $-c$. Operations between the Euclidean tangent space $T_{\mathbf{0}}\mathbb{D}_c^d$ (approximated as $\mathbb{R}^d$ at the origin) and the hyperbolic manifold are mediated by exponential ($\exp_{\mathbf{0}}^c$) and logarithmic ($\log_{\mathbf{0}}^c$) maps. The induced Poincaré distance $d_{\mathbb{D}}(\boldsymbol{z}_i, \boldsymbol{z}_j)$ grows exponentially as points approach the boundary, providing the necessary capacity to embed hierarchies without distortion.

\subsection{Constructing the Cognitive Geometric Prior}
The goal of this phase is to train the Hyperbolic Cognitive Network (HCN) as a "geometric teacher" that effectively disentangles cognitive states.

\paragraph{Hyperbolic Disentanglement.} Given an input $X$, an encoder extracts the hidden state $\boldsymbol{h}_{enc}$. To decouple the four cognitive dimensions, we project dimension-specific base vectors into the hyperbolic space: $\boldsymbol{z}_k = \exp_{\mathbf{0}}^c(\boldsymbol{W}_k \boldsymbol{h}_{enc})$ for $k \in \{emo, thk, stn, int\}$. These features are processed and aggregated via cross-dimensional attention in the tangent space to form a unified \textit{cognitive anchor} $\boldsymbol{v}_{cog}$.

\paragraph{Geometry-Aware Optimization.} To enforce structural separation, we optimize a composite objective $\mathcal{L}_{HCN} = \mathcal{L}_{task} + \mathcal{L}_{hyp}$. Here, $\mathcal{L}_{task}$ is the multi-task cross-entropy loss for classification. Crucially, $\mathcal{L}_{hyp}$ is a hyperbolic geometric regularization loss based on the Poincaré distance. It plays the role of contrastive learning, bringing samples with the same cognitive features closer, while pushing away those with different cognitive features:
\begin{equation}
\begin{aligned}
\mathcal{L}_{hyp} &= \sum_{i,j} \left[ \mathbb{I}_{y_i = y_j}d_{\mathbb{D}}(\boldsymbol{z}_i, \boldsymbol{z}_j)^2 \right. \\
&\quad \left. + \mathbb{I}_{y_i \neq y_j}\max(0, m - d_{\mathbb{D}}(\boldsymbol{z}_i, \boldsymbol{z}_j))^2 \right]
\end{aligned}
\end{equation}
where $m$ is the margin controlling the separation of distinct classes. This ensures the learned feature space reflects the intrinsic hierarchy of the data, resolving the crowding issue.

\subsection{Hyperbolic-Guided Alignment Tuning}

Hyperbolic-Guided Alignment Tuning (HGAT) can effectively disentangle cognitive states, and we need to inject this geometric prior knowledge into LLMs. Considering the inherent differences between the Euclidean space used by LLMs and hyperbolic space, we design HGAT to align these two topological structures.

\paragraph{Soft Prompt Injection.} We construct a Cognitive Projection Network $\mathcal{P}_\theta$ that maps the static anchor $\boldsymbol{v}_{cog}$ from the HCN to a sequence of dynamic soft prompt embeddings $\boldsymbol{E}_{prior} \in \mathbb{R}^{L \times d_{model}}$, where $L$ is the prompt length. These embeddings are prepended to the input embeddings to guide the generation process:
\begin{equation}
    \small
    \boldsymbol{E}_{prior} = \mathcal{P}_\theta(\boldsymbol{v}_{cog}) = a \cdot \tanh\left(\boldsymbol{W}_2 \cdot \sigma(\operatorname{LN}(\boldsymbol{W}_1 \boldsymbol{v}_{cog}))\right),
\end{equation}
\begin{equation}
    \widetilde{\boldsymbol{X}} = \boldsymbol{E}_{prior} \oplus \operatorname{Emb}(C) \oplus \operatorname{Emb}(T),
\end{equation}
where $\boldsymbol{W}_1, \boldsymbol{W}_2$ are projection weights, $\operatorname{LN}$ denotes LayerNorm applied to the projected features before the activation $\sigma$ (consistent with Figure~\ref{fig:framework}), and $a$ is a scaling factor bounding the soft prompt values. Here $C$ denotes the \emph{Context} (e.g., the source tweet or topic information that frames a user's speech) and $T$ denotes the \emph{Target} post (the user's actual speech that we aim to analyze). $\widetilde{\boldsymbol{X}}$ is the model input formed by concatenating the soft prompt with the context and target embeddings; $\oplus$ denotes concatenation along the sequence length dimension, ensuring the soft prompts serve as a prefix.

\paragraph{Semantic-Cognitive Topology Loss.} 
To ensure the generated semantic representations remain topologically consistent with the intended cognitive goals, we propose the Semantic-Cognitive Topology (SCT) Loss. We extract the hidden state of the last token in the LLM's generation sequence, $\boldsymbol{h}_{last}$. An alignment projector $A_\phi$ maps this state back to the tangent space to obtain $\hat{\boldsymbol{v}}_{sem}$, which is then constrained to align with the cognitive anchor $\boldsymbol{v}_{cog}$ using cosine similarity:
\begin{equation}
    \mathcal{L}_{SCT} = 1 - \frac{\hat{\boldsymbol{v}}_{sem}^T \cdot \boldsymbol{v}_{cog}}{\|\hat{\boldsymbol{v}}_{sem}\|_2 \|\boldsymbol{v}_{cog}\|_2 + \epsilon}.
\end{equation}

Since the Poincaré ball model is conformal, the angular distance between vectors in the tangent space $T_\mathbf{0}\mathbb{D}_c^d$ preserves the angular relations of geodesics emanating from the origin, ensuring that semantic directionality is maintained. $\mathcal{L}_{SCT}$ mathematically enforces a local isomorphism between the LLM's semantic space and the HCN's cognitive tangent space, penalizing any semantic drift that deviates from the predetermined cognitive hierarchy. The final joint optimization objective combines the $\mathcal{L}_{SFT}$ and $\mathcal{L}_{SCT}$:
\begin{equation}
    \mathcal{L}_{total} = \underbrace{-\sum_{t=1}^{|Y|} \log P(y_t | \widetilde{\boldsymbol{X}}, y_{<t})}_{\mathcal{L}_{SFT}} + \lambda \cdot \mathcal{L}_{SCT},
\end{equation}
where $\mathcal{L}_{SFT}$ represents the standard auto-regressive generation loss, $\lambda$ is a hyperparameter balancing generation quality and structural constraints. The $\mathcal{L}_{SCT}$ loss incorporates geometric constraints into the LoRA parameters of the LLM. Consequently, there is no need to use HCN or $\boldsymbol{E}_{\textit{prior}}$ during inference, and compared with general LLMs, HyCoLLM exhibits no additional inference latency.

\subsection{Geometric Capacity Analysis}
\label{sec:theory_overview}

We attribute cognitive crowding to a fundamental geometric mismatch between hierarchical cognitive structures and Euclidean representations. Empirically, the labels in CognitiveBench exhibit strong hierarchy, as indicated by consistently low Gromov $\delta$ values. This motivates a theoretical analysis of the embedding capacity required to represent such structures.

\textbf{Proposition 1 (Geometric Capacity Mismatch).}
\textit{Let $\mathcal{T}$ be a hierarchical taxonomy with branching factor $b>1$ and depth $k$. Embedding $\mathcal{T}$ into a $d$-dimensional Euclidean space $\mathbb{R}^d$ with bounded distortion requires the embedding radius to grow exponentially with $k$. In contrast, a $d$-dimensional hyperbolic space admits a low-distortion embedding with radius growing only linearly in $k$.}

This result shows that the polynomial volume growth of Euclidean space is insufficient to accommodate the exponential expansion of hierarchical cognitive structures, inevitably leading to crowding. Hyperbolic geometry, with exponential volume growth, provides a principled solution by enabling faithful separation of hierarchically related cognitive states. A formal proof is given in Appendix~\ref{app:theory}.

\section{Experiment}

\paragraph{Baselines.}
We compare our method against three groups of baselines: (1) \textbf{Human Experts}: Providing an upper-bound reference for deep cognitive understanding; (2) \textbf{Closed-Source LLMs}: Including GPT-4o~\citep{achiam2023gpt}, Gemini2.5~\citep{comanici2025gemini}, etc., evaluated under Few-Shot and Chain-of-Thought (CoT) settings; (3) \textbf{Open-Source LLMs}: Including Llama-3.1-8B~\citep{touvron2023llama}, Qwen3-8B~\citep{yang2025qwen3}, etc., fine-tuned with Supervised Fine-Tuning (SFT) on the CognitiveBench.

\paragraph{Metrics.}
Beyond standard \textbf{Accuracy (ACC)} and \textbf{Macro-F1} for individual dimensions, we introduce holistic metrics to evaluate the synergy of cognitive state prediction: \textbf{Partial Match Accuracy@\textit{k} (PMA@\textit{k})} measures the proportion of samples where the model correctly predicts \textit{at least k} dimensions. Specifically, \textbf{PMA@4} serves as the strictest metric (Exact Match), requiring correct predictions for emotion, thinking, stance, and intent simultaneously. This metric captures the "overall cognitive state" consistency. \textbf{\textit{Hamming loss} ($\downarrow$)}: The proportion of incorrectly predicted labels relative to the total number of labels.

\paragraph{Implementation Details.}
We train HyCoLLM using the \texttt{trl} framework~\citep{vonwerra2022trl}, along with 4-bit quantization and LoRA adapters. The training process runs for 3 epochs with a learning rate set to 2e-4, and the weight of the SCT loss is fixed at $\lambda=1.0$ based on sensitivity analysis. During inference, the model is deployed using \texttt{vllm}~\citep{kwon2023efficient}, with a temperature of 0 and a max\_token of 128. Both the baselines and HyCoLLM use identical settings during training and inference to ensure fair comparison. For more settings, please refer to Appendix~\ref{app:exp_details}.

\subsection{Main Results}

\begin{table*}[tb]
\centering
\Large
\resizebox{\textwidth}{!}{%
\begin{tabular}{ll | cc cc cc cc | ccc | c }
\toprule
\multicolumn{2}{c|}{\multirow{2}{*}{Model}} & 
\multicolumn{2}{c}{Emotion} & 
\multicolumn{2}{c}{Thinking} & 
\multicolumn{2}{c}{Stance} & 
\multicolumn{2}{c|}{Intent} & 
\multirow{2}{*}{PMA@2} & 
\multirow{2}{*}{PMA@3} & 
\multirow{2}{*}{PMA@4} & 
\multirow{2}{*}{Hamming Loss $\downarrow$} \\
\multicolumn{2}{c|}{} & ACC & F1 & ACC & F1 & ACC & F1 & ACC & F1 & & & & \\ 
\midrule

\rowcolor{header_blue} \multicolumn{14}{c}{\textbf{Human}} \\
\multicolumn{2}{c|}{Averaged from 3 experts} & 0.7251 & 0.5348 & 0.6400 & 0.4816 & 0.8021 & 0.6713 & 0.7201 & 0.5762 & 0.9092 & 0.6836 & 0.3076 & 0.0797 \\
\midrule
\rowcolor{header_blue} \multicolumn{14}{c}{\textbf{Close-Source}} \\
\multirow{4}{*}{\shortstack[l]{Small-size\\Few-Shot \& CoT}} 
 & GPT4o-mini & 0.4519 & 0.3015 & 0.3060 & 0.2099 & 0.5466 & 0.5470 & 0.5541 & 0.4181 & 0.6436 & 0.2695 & 0.0442 & 0.1676 \\
 & Gemini2.5-Flash-Lite & 0.5382 & 0.3012 & 0.3229 & 0.2532 & 0.5554 & 0.5501 & 0.5750 & 0.4242 & 0.6814 & 0.3101 & 0.0618 & 0.1573 \\
 & Grok-4-fast & 0.4266 & 0.2904 & 0.3066 & 0.2516 & 0.5329 & 0.4967 & 0.5556 & 0.5153 & 0.6361 & 0.2499 & 0.0183 & 0.1702 \\
 & Qwen-Turbo & 0.4543 & 0.2641 & 0.3335 & 0.2277 & 0.4873 & 0.4248 & 0.5528 & 0.4002 & 0.6145 & 0.2485 & 0.0504 & 0.1694 \\ 
\cmidrule{1-14}
\multirow{4}{*}{\shortstack[l]{Large-size\\Few-Shot \& CoT}} 
 & GPT-4o & 0.5156 & 0.2983 & 0.3259 & 0.2596 & \underline{0.6596} & \underline{0.6430} & 0.4834 & 0.4215 & 0.6870 & 0.2994 & 0.0569 & 0.1572 \\
 & DeepSeek-V3.1 & 0.5293 & 0.3282 & 0.3196 & 0.2639 & 0.5713 & 0.5622 & 0.5327 & 0.4258 & 0.6719 & 0.2917 & 0.0588 & 0.1597 \\
 & Claude-haiku-4.5 & 0.4424 & 0.3157 & 0.3024 & 0.2387 & 0.5234 & 0.5046 & 0.5459 & 0.3610 & 0.6375 & 0.2359 & 0.0335 & 0.1708 \\
 & Qwen-Max & 0.4531 & 0.3094 & 0.2689 & 0.2056 & 0.5780 & 0.5205 & 0.5630 & 0.4185 & 0.6489 & 0.2782 & 0.0286 & 0.1662 \\
\midrule
\rowcolor{header_blue} \multicolumn{14}{c}{\textbf{Open-Source}} \\
\multirow{5}{*}{After SFT} 
 & Llama3.1-8b & 0.5270 & 0.3256 & 0.3703 & 0.1763 & 0.6392 & 0.5543 & 0.5366 & 0.3351 & 0.7047 & 0.3480 & 0.0795 & 0.1516 \\
 & Qwen3-8B & 0.5867 & 0.3228 & 0.4308 & 0.3099 & 0.6488 & 0.5332 & 0.6113 & 0.4995 & 0.7987 & 0.4001 & 0.1091 & 0.1349 \\
 & Qwen2.5-7B & 0.5416 & 0.3244 & 0.4876 & 0.2381 & 0.6207 & 0.5938 & 0.6053 & 0.4444 & 0.7610 & 0.4240 & 0.1101 & 0.1286 \\
 & DeepSeek-R1-Qwen3 & 0.5286 & 0.2874 & 0.3378 & 0.2274 & 0.5755 & 0.4802 & 0.5421 & 0.4285 & 0.6665 & 0.3007 & 0.0639 & 0.1517 \\
 & Mistral-7b & 0.3526 & 0.1202 & 0.1978 & 0.0850 & 0.4402 & 0.3287 & 0.2530 & 0.1192 & 0.2741 & 0.1815 & 0.0135 & 0.2049 \\
\midrule
\rowcolor{header_blue} \multicolumn{14}{c}{\textbf{Ours}} \\
\multirow{2}{*}{After SFT} 
 & HyCoLLM-Qwen3-8b & \textbf{0.6286} & \textbf{0.3505} & \textbf{0.5334} & \textbf{0.3867} & \textbf{0.7033} & \textbf{0.6585} & \underline{0.6475} & \textbf{0.5473} & \textbf{0.8439} & \textbf{0.5359} & \textbf{0.1549} & \textbf{0.1156} \\
 & HyCoLLM-Llama3.1-8b & \underline{0.6133} & \underline{0.3289} & \underline{0.5091} & \underline{0.3771} & 0.6508 & 0.5543 & \textbf{0.6584} & \underline{0.5392} & \underline{0.8179} & \underline{0.5047} & \underline{0.1431} & \underline{0.1215} \\ 
\bottomrule
\end{tabular}%
}
\caption{Overall averaged performance across three datasets (CUT, UE, and DEI). The best results are highlighted in \textbf{bold}, and the second-best are \underline{underlined}. For Hamming Loss, lower is better ($\downarrow$).}
\label{tab:average_results}
\end{table*}

Table~\ref{tab:average_results} reports the averaged performance across CUT, UE and DEI datasets, with detailed sub-dataset results provided in Appendix~\ref{app:sub_main_results}. The results highlight clear limitations in the holistic cognitive understanding of current LLMs and demonstrate the effectiveness of our approach.

Despite strong reasoning capabilities, baseline models exhibit pronounced cognitive crowding. Even GPT-4o with CoT prompting attains only 5.69\% \textit{PMA@4} accuracy, far below the human benchmark of 30.76\%. This gap indicates that existing LLMs can often predict individual dimensions, such as stance, but fail to maintain consistency across emotion, thinking, stance, and intent. Open-source models fine-tuned with standard LoRA-based SFT show moderate gains, with Qwen3-8B reaching 10.91\% \textit{PMA@4}, but still incur high \textit{Hamming Loss}. We further find that zero-shot and few-shot Chain-of-Thought (CoT) prompting alone is insufficient and can even degrade structured multi-dimensional predictions, as LLMs frequently fail to adhere to the strict 4-dimensional schema (see Appendix~\ref{app:cot_baselines}). More importantly, a scaling-law analysis across 1.7B, 8B, and 32B backbones (Appendix~\ref{app:scaling}) reveals a striking performance inversion: even Qwen3-32B under standard SFT (PMA@4 = 9.42\%) underperforms our \emph{8B} HyCoLLM (13.96\%), which directly rules out ``task difficulty'' or insufficient parameter capacity as the root cause and instead confirms that the bottleneck is geometric. These results together support our hypothesis that Euclidean representations entangle semantically related dimensions, leading to systematic cognitive crowding.

\begin{figure}[t]
    \centering
    \includegraphics[width=1.0\linewidth]{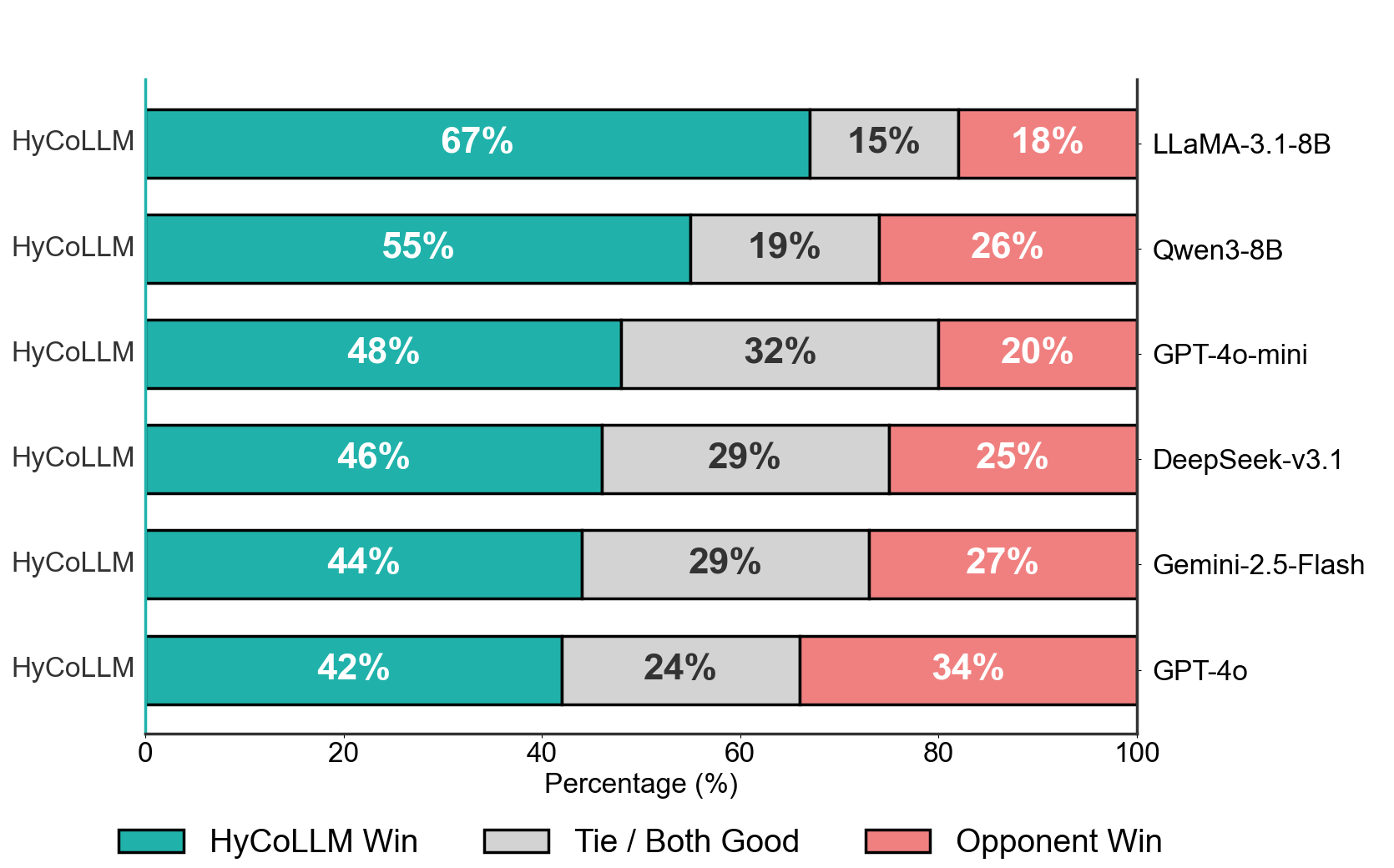}
    \caption{Results of pairwise human evaluations on 100 randomly sampled instances. HyCoLLM here refers to the HyCoLLM-Qwen3-8B variant. Evaluators judged the overall quality and logical self-consistency of the predicted 4-dimensional cognitive states.} 
    \label{fig:human_eval}
\end{figure}

HyCoLLM substantially alleviates this issue. With Qwen3-8B, HyCoLLM achieves the strongest overall performance among automated evaluations, improving \textit{PMA@4} accuracy to 15.49\%, nearly three times that of GPT-4o and over 40\% higher than the best SFT baseline. Gains are particularly pronounced in difficult dimensions such as Thinking and intent. HyCoLLM also yields the lowest \textit{Hamming Loss} of 0.1156, narrowing the gap to human experts (0.0797). These results indicate that modeling cognitive structure in hyperbolic space enables small LLMs to surpass much larger LLMs.

\paragraph{Human Evaluation.}

To complement automatic metrics, we conducted a blind pairwise human evaluation to assess the holistic consistency of the predicted cognitive states (emotion, thinking, stance, and intent). As shown in Figure \ref{fig:human_eval}, HyCoLLM consistently outperforms comparative models. Notably, it achieves a 67\% win rate against the Llama3.1-8b after SFT, validating the necessity of geometric alignment. Furthermore, HyCoLLM surpasses the proprietary model GPT-4o (42\% wins vs. 34\% losses), suggesting that a specialized hyperbolic geometric prior captures the nuanced hierarchy of human cognition more effectively than merely scaling model parameters. The result in the pairwise human evaluation represents the average values evaluated by three experts. For the evaluation manual, please refer to Figure~\ref{fig:eval_prompt}.

\begin{table*}[t]
\centering

\resizebox{\textwidth}{!}{
\begin{tabular}{l | cc cc cc cc | ccc | c }
\toprule
\multirow{2}{*}{Method} & \multicolumn{2}{c}{Emotion} & \multicolumn{2}{c}{Thinking} & \multicolumn{2}{c}{Stance} & \multicolumn{2}{c|}{intent} & \multirow{2}{*}{PMA@2} & \multirow{2}{*}{PMA@3} & \multirow{2}{*}{PMA@4} & \multirow{2}{*}{H-Loss $\downarrow$} \\
 & ACC & F1 & ACC & F1 & ACC & F1 & ACC & F1 & & & & \\ 
\midrule
\textbf{HyCoLLM (Full)} & \textbf{0.6286} & 0.3505 & \textbf{0.5334} & \textbf{0.3867} & \textbf{0.7033} & \textbf{0.6585} & \textbf{0.6475} & \textbf{0.5473} & \textbf{0.8439} & \textbf{0.5359} & \textbf{0.1549} & \textbf{0.1156} \\
\midrule
w/o Hyperbolic & 0.6241 & 0.3175 & 0.4535 & 0.3246 & 0.6770 & 0.5992 & 0.6323 & 0.4760 & 0.8187 & 0.4829 & 0.1216 & 0.1257 \\
w/o Geo-Regularization & 0.6179 & 0.3516 & 0.4517 & 0.3260 & 0.6693 & 0.5861 & 0.6303 & 0.4980 & 0.8073 & 0.4714 & 0.1204 & 0.1275 \\
w/o Cross-Interaction & 0.6174 & 0.2898 & 0.4699 & 0.3417 & 0.6711 & 0.5910 & 0.6286 & 0.4810 & 0.8091 & 0.4884 & 0.1165 & 0.1260 \\
\midrule
w/o Soft Prompt & 0.6164 & 0.2835 & 0.4536 & 0.3293 & 0.6561 & 0.5800 & 0.6335 & 0.4782 & 0.7998 & 0.4711 & 0.1176 & 0.1281 \\
w/o SCT Loss & 0.6228 & \textbf{0.3530} & 0.4795 & 0.3526 & 0.6852 & 0.6225 & 0.6243 & 0.4818 & 0.8149 & 0.4881 & 0.1252 & 0.1249 \\
w/o Cognitive Anchor & 0.6174 & 0.2859 & 0.4641 & 0.3153 & 0.6646 & 0.5877 & 0.6372 & 0.4782 & 0.8099 & 0.4788 & 0.1215 & 0.1261 \\
\bottomrule
\end{tabular}
}
\caption{Ablation analysis averaged across three datasets. We investigate the contribution of each component in the HCN and HGAT phases.}
\label{tab:ablation_avg}
\end{table*}

\subsection{Ablation Study}

Table~\ref{tab:ablation_avg} summarizes the ablation results, with detailed analyses in Appendix~\ref{app:ablation_sub}. Removing the hyperbolic geometry and reverting to Euclidean space leads to a sharp drop in \textit{PMA@4} accuracy, from 15.49\% to 12.16\%, despite retaining the same architecture and objectives. This confirms that Euclidean representations exacerbate cognitive crowding and undermine holistic consistency. Removing geometric regularization increases \textit{Hamming Loss}, indicating that explicit topological constraints are necessary to preserve fine-grained cognitive distinctions. Removing cross-dimensional interaction primarily degrades thinking and stance performance, highlighting the intrinsic coupling among cognitive dimensions. In the alignment phase, both soft prompts and the Semantic Cognitive Topology loss are critical. Removing soft prompts, \textit{PMA@4} accuracy drops to 11.76\%, while removing the alignment loss also degrades performance. Finally, removing the cognitive anchor weakens overall consistency, confirming the importance of holistic cognitive representation for multi-dimensional reasoning. A complementary two-path cumulative ablation, which successively removes the Soft Prompt and the SCT Loss in either order, is reported in Appendix~\ref{app:cumulative_ablation}: regardless of order the removal of either module causes substantial degradation (15.49\% $\to$ 12.52\%/11.76\% $\to$ 10.91\%), and the Soft Prompt consistently shows a stronger standalone effect, confirming the synergistic contribution of the two modules.

\begin{figure}[t]
    \centering
    \vspace{-10pt}
    \includegraphics[width=0.98\linewidth]{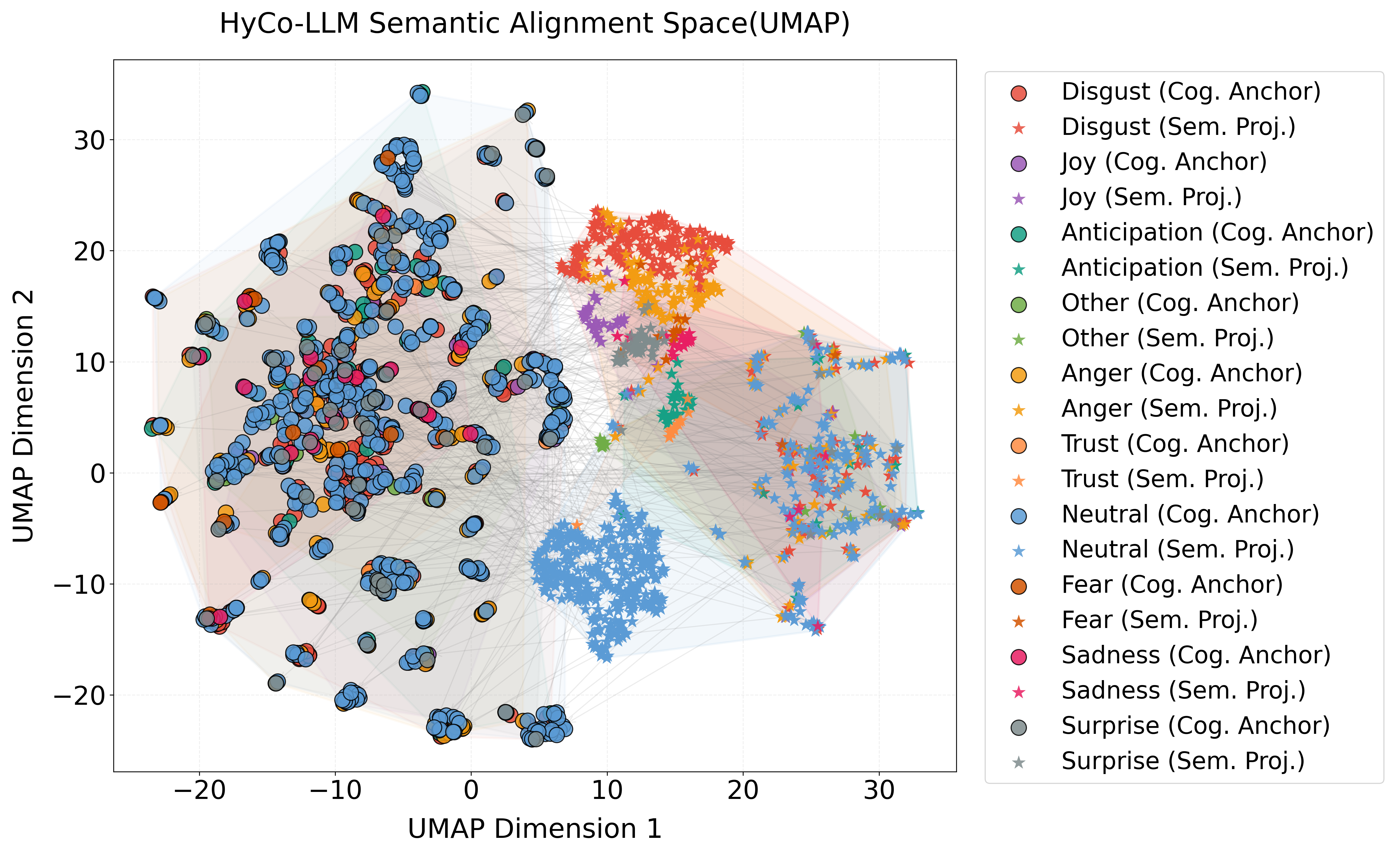}
    \caption{Visualization of the Semantic-Cognitive Alignment using UMAP. The Circles ($\circ$) represent the cognitive anchors derived from the HCN, while the Stars ($\star$) denote the LLM's semantic features.}
    \label{fig:umap_vis}
    \vspace{-10pt}
\end{figure}

\subsection{Qualitative Analysis}

To intuitively verify the impact of HyCoLLM, we visualize the alignment between the cognitive geometric prior and the LLM's semantic space using UMAP (Figure \ref{fig:umap_vis}). Cognitive anchors form well-separated clusters corresponding to distinct cognitive states, while semantic features are drawn towards their associated anchors. This indicates that HyCoLLM achieves an optimal balance: it injects structural cognitive constraints to rectify the crowding in Euclidean space, while retaining the necessary semantic richness required for language generation. The anchors thus act as "semantic attractors," imposing a hierarchical order on the LLM's representation space without stripping away linguistic nuance.

\subsection{Generalization to Unseen Topics}
To assess robustness, we evaluate models trained on CUT against the unseen FRIR dataset (See Appendix~\ref{app: generalization} for details). As shown in Table~\ref{tab:frir_generalization_concise}, HyCoLLM demonstrates superior cross-domain adaptability, achieving the lowest \textit{Hamming Loss} (12.5\%) and leading PMA@4 accuracy (11.2\%), significantly outperforming other baselines. This confirms that the hyperbolic prior effectively disentangles intrinsic cognitive states across domains. However, the \textit{Stance} dimension shows limited transferability, highlighting a boundary condition when targets are tied to surface semantics. 
\begin{table}[tb]
\centering
\Large
\vspace{-10pt}
\resizebox{\columnwidth}{!}{%
\begin{tabular}{l|cccc|cccc}
\toprule
\multicolumn{1}{c|}{Model} & Emo. & Thk. & Stn. & Int. & PMA@2 & PMA@3 & PMA@4 & H-Loss$\downarrow$ \\
\midrule
Human & 57.3 & 51.1 & 39.7 & 64.6 & 92.7 & 71.8 & 33.3 & 6.9 \\
\midrule
Gemini2.5-Flash & 36.6 & 25.1 & \textbf{49.7} & 47.6 & 68.9 & 33.9 & 6.2 & 14.6 \\
    GPT-4o & \underline{44.5} & 21.0 & 42.7 & 37.9 & 60.1 & 21.2 & 3.6 & 16.7 \\
\midrule
Qwen3-8B & 33.6 & \textbf{34.8} & 33.1 & 39.1 & 72.6 & 38.2 & 9.2 & 13.8 \\
Llama3.1-8b & 20.7 & 20.4 & 27.3 & 32.8 & 61.2 & 24.9 & 5.3 & 15.9 \\
\midrule
HyCoLLM-Llama3.1 & \textbf{45.5} & \underline{33.7} & 18.7 & \textbf{61.5} & \textbf{77.4} & \textbf{42.7} & \textbf{11.2} & \textbf{12.5} \\
HyCoLLM-Qwen3 & 37.3 & 32.9 & \underline{31.6} & \underline{50.4} & \underline{74.6} & \underline{38.3} & \underline{10.6} & \underline{13.5} \\
\bottomrule
\end{tabular}%
}
\caption{Generalization results (\%) on FRIR (trained on CUT).}
\label{tab:frir_generalization_concise}
\vspace{-10pt}
\end{table}

\subsection{Supplementary Analysis}
We provide additional analyses to examine the practicality and robustness of HyCoLLM. Specifically, we show that HyCoLLM incurs only a modest training overhead while introducing \textit{zero additional inference latency} (Appendix~\ref{app:efficiency}), exhibits \textit{stable performance across a wide range of hyperparameters} with a clear optimum (Appendix~\ref{app:hyperparams}), and \textit{consistently scales across model sizes} from 1.7B to 32B parameters (Appendix~\ref{app:scaling}). Furthermore, detailed comparative and ablation studies on multiple sub-datasets demonstrate that the proposed geometric prior yields \textit{robust improvements across diverse cognitive domains} (Appendix~\ref{app:sub_main_results}).

\section{Conclusion}
We construct CognitiveBench, a benchmark for joint modeling of emotion, thinking, stance, and intent, and show that current LLMs struggle with consistent multi-dimensional cognitive understanding. To address this limitation, we propose HyCoLLM, which aligns LLM representations with a structured cognitive manifold through hyperbolic geometry and geometry-aware tuning. Extensive experiments show that HyCoLLM significantly mitigates cognitive crowding and achieves state-of-the-art performance on CognitiveBench. 

\section{Limitation}

While HyCoLLM demonstrates clear advantages in multi-dimensional cognitive modeling, several directions remain open for future exploration. First, although most geometric operations are approximated in the tangent space, training still involves Riemannian optimization on hyperbolic manifolds, including exponential and logarithmic mappings. Future work may investigate more efficient approximations or alternative optimization schemes that further reduce computational overhead, facilitating deployment in large-scale or latency-sensitive settings.

Second, the construction of the cognitive geometric prior relies on high-quality, fine-grained annotations such as those provided by CognitiveBench. Extending this framework to low-resource languages or domains without rich psychological annotations remains an important direction. Promising avenues include weakly supervised learning, cross-lingual transfer, or leveraging self-supervised signals to approximate cognitive structure with reduced annotation cost.

Finally, although soft prompts are adopted to limit interference with the base language model, the broader impact of geometry-guided alignment on general-purpose capabilities, such as code generation or mathematical reasoning, warrants further investigation. Future studies could explore adaptive or task-aware alignment strategies that preserve cognitive consistency while maintaining strong performance on purely logical tasks.

\section{Ethical Considerations}

By enabling deeper inference of human cognitive states, including latent intents and thinking patterns, HyCoLLM has the potential to improve human–computer interaction and support more adaptive language understanding. At the same time, such capabilities raise important ethical considerations, which we have actively addressed throughout data collection, model design, and deployment recommendations.

A primary concern is the risk of misuse. Models with enhanced theory-of-mind-like abilities could, if misused, generate persuasive or manipulative content that targets individual psychological vulnerabilities. To reduce this risk, we restrict the release of model weights under controlled licenses and advocate for the use of application-level safety mechanisms, such as content moderation, usage monitoring, and access control, when deploying HyCoLLM in real-world systems. These measures aim to limit malicious exploitation while preserving legitimate research and application use.

Privacy protection is another critical consideration. Since HyCoLLM can infer emotional and cognitive patterns from text, there is a risk of unintentionally exposing sensitive personal information. To mitigate this, all data used in CognitiveBench were sourced from public platforms and processed with strict anonymization procedures, including the removal of user identifiers and personally identifiable information. We further emphasize that downstream applications, particularly in sensitive domains such as mental health support, should comply with data protection regulations, obtain explicit user consent, and avoid over-interpretation of inferred cognitive states.

Finally, since HyCoLLM encodes expert-annotated cognitive structures as geometric priors, biases present in the annotation process may be preserved or amplified. To address this risk, we employed multiple annotators with relevant domain expertise and enforced consensus-based labeling to reduce individual bias. Looking ahead, we encourage future expansions of CognitiveBench to incorporate more diverse populations and cultural contexts, as well as the development of systematic bias detection and mitigation techniques tailored to structured cognitive representations. These combined efforts reflect our commitment to minimizing ethical risks while advancing research on cognitive alignment in large language models.

\section*{Acknowledgments}
This work was supported in part by the National Key Research and Development Program of China under Grant 2023YFB3107000. This work has also been supported by the New Cornerstone Science Foundation through the XPLORER PRIZE.


\bibliography{custom}

\appendix

\section{Construction Details of CognitiveBench}
\label{app:annotation}

\subsection{Theoretical Framework and Label Definitions}
\label{app:definitions}

To ensure the validity and psychological depth of \textit{CognitiveBench}, we grounded our taxonomy in established theories from cognitive science, psychology, and linguistics. This section details the theoretical rationale behind each dimension and the definitions of the corresponding labels.

\paragraph{Emotion Dimension: Plutchik's Evolutionary Model.}
We adopted \textit{Plutchik's Wheel of Emotions}~\citep{plutchik1980emotion} as our foundational framework. Unlike simple emotion analysis (positive/negative), Plutchik's model provides a biologically grounded classification of emotions based on evolutionary survival functions (e.g., Fear $\rightarrow$ Escape, Anger $\rightarrow$ Attack). This granularity is essential for understanding the functional role of emotions in social media discourse—for instance, distinguishing whether a user is debating out of "Anger" (hostility) or "Disgust" (rejection). We categorize the 8 basic emotions into three valences—\textit{Positive}, \textit{Negative}, and \textit{Neutral}—to align with the hierarchical structure of other cognitive dimensions.

\begin{table*}[htpb]
\centering

\resizebox{0.95\textwidth}{!}{
\begin{tabularx}{\textwidth}{@{}l l X@{}}
\toprule
\textbf{Category} & \textbf{Label} & \textbf{Definition} \\ \midrule
\multirow{4}{*}{\textbf{Positive}} 
 & \textbf{Joy} & A feeling of great pleasure and happiness (Functional role: Gain/Possession). \\
 & \textbf{Trust} & Reliance on the integrity or strength of a person/thing (Functional role: Acceptance). \\
 & \textbf{Anticipation} & A state of looking forward to a future event (Functional role: Exploration). \\ 
 & \textbf{Surprise} & A feeling of shock caused by the unexpected (Note: Can be valenced, categorized here as mild astonishment). \\ \midrule
\multirow{4}{*}{\textbf{Negative}} 
 & \textbf{Anger} & A strong feeling of hostility due to frustration/injustice (Functional role: Destruction/Attack). \\
 & \textbf{Disgust} & A feeling of revulsion or strong disapproval (Functional role: Rejection/Expulsion). \\
 & \textbf{Fear} & An intense emotion triggered by perceived threat (Functional role: Protection/Escape). \\
 & \textbf{Sadness} & An unpleasant state typically caused by loss (Functional role: Reintegration). \\ \midrule
\textbf{Neutral} 
 & \textbf{Neutral} & An objective state not showing clear emotional tendency (Functional role: Observation). \\ 
 \bottomrule
\end{tabularx}
}
\caption{Taxonomy of Emotion Labels (Based on Plutchik's Evolutionary Model)}
\label{tab:def_emotion}
\end{table*}

\paragraph{Thinking Dimension: Dual-Process Theory.}
We leverage \textit{Dual-Process Theory}\citep{kahneman2011thinking} to decode the cognitive depth of user posts. This theory distinguishes between \textit{System 1} (Intuitive, fast, heuristic-based) and \textit{System 2} (Analytical, slow, rule-based).
Standard text classification often ignores this distinction. By explicitly modeling "Thinking Style," we can differentiate between a user who supports a stance via \textit{Logical Deduction} (System 2) versus one who relies on \textit{Identity-driven Conformity} (System 1). This is crucial for identifying the robustness of public opinion. We further granularize these systems into specific cognitive values derived from \textit{Social Identity Theory}~\citep{turner1986significance} and \textit{Heuristics research}.

\begin{table*}[h]
\centering

\resizebox{0.95\textwidth}{!}{%
\begin{tabularx}{\textwidth}{@{}l l X@{}}
\toprule
\textbf{Category} & \textbf{Label} & \textbf{Theoretical Basis \& Definition} \\ \midrule
\multirow{4}{*}{\textbf{Intuitive}} 
 & \textbf{Subjective Evaluation} & \textit{Overconfidence Effect:} Irrational confidence stemming from a belief of full understanding without verification. \\
 & \textbf{Identity Conformity} & \textit{Social Identity Theory:} Adopting values or attitudes primarily to signal membership in a specific group or allegiance to authority. \\
 & \textbf{Emotional Judgment} & \textit{Affect Heuristic:} Relying on current emotional states (e.g., "I feel bad about this") as valid evidence for judgment. \\
 & \textbf{Experience-based} & \textit{Availability Heuristic:} Basing judgments on immediate, personal examples that come to mind easily, rather than statistical data. \\ \midrule
\multirow{4}{*}{\textbf{Analytical}} 
 & \textbf{Logical} & \textit{Deductive Reasoning:} Systematically deriving conclusions from premises using valid logical structures. \\
 & \textbf{Balanced Consideration} & \textit{Rational Choice Theory:} Explicitly weighing pros and cons or comparing utilities before forming a preference. \\
 & \textbf{Evidence-based} & \textit{Information Integration:} Consciously citing and evaluating external facts, data, or sources to support a claim. \\
 & \textbf{Critical} & \textit{Critical Thinking:} Reflectively evaluating one's own assumptions or questioning the validity of external information. \\ \bottomrule
\end{tabularx}%
}
\caption{Taxonomy of Thinking Styles and Values (Based on Dual-Process Theory)}
\label{tab:def_thinking}
\end{table*}

\paragraph{Intent Dimension: Speech Act Theory.}
To capture the communicative goal behind a post, we employ \textit{Speech Act Theory}~\citep{austin1975things, searle1976classification}. This theory posits that language is not just used to describe the world, but to perform actions (illocutionary acts).
By mapping posts to categories like \textit{Representatives} (stating facts), \textit{Directives} (requesting action), and \textit{Expressives} (revealing state), we move beyond topic classification to understand user intent. For example, distinguishing between a user who is "Seeking Information" (Directive) versus one who is "Expressing Conflict" (Expressive) allows the AI to generate more appropriate and empathetic responses.

\begin{table*}[h]
\centering

\resizebox{0.95\textwidth}{!}{%
\begin{tabularx}{\textwidth}{@{}l l X@{}}
\toprule
\textbf{Category} & \textbf{Label} & \textbf{Definition} \\ \midrule
\multirow{2}{*}{\textbf{Representatives}} & \textbf{Info. Sharing} & Providing objective facts, data, or personal experiences to convey information. \\
 & \textbf{Opinion Exp.} & Expressing subjective views, judgments, or positions on a topic. \\ \midrule
\multirow{2}{*}{\textbf{Directives}} & \textbf{Info. Seeking} & Attempting to obtain information or understand others' views through questions. \\
 & \textbf{Call to Action} & Explicitly instructing, urging, or persuading others to take specific offline or online actions. \\ \midrule
\multirow{3}{*}{\textbf{Expressives}} & \textbf{Connection} & Expressing positive social attitudes, such as support, gratitude, or identification with others. \\
 & \textbf{Conflict} & Expressing disagreement, criticism, sarcasm, or negation towards others' views. \\
 & \textbf{Emotional Exp.} & Directly revealing the speaker's internal emotional state (catharsis) without necessarily targeting others. \\ \bottomrule
\end{tabularx}%
}
\caption{Taxonomy of  Labels (Based on Speech Act Theory)}
\label{tab:def_intent}
\end{table*}

\paragraph{Stance Dimension.}
Stance annotation captures the user's position relative to a specific target. Grounded in \textit{Social Judgment Theory}~\citep{sherif1961social}, we define the stance not just as binary support/oppose, but consider the "Latitude of Non-Commitment" (Unclear/Neutral).
\begin{itemize}
    \item \textbf{CUT (Trade War):} \textit{Support US}, \textit{Support China}, or \textit{Unclear}.
    \item \textbf{UE (US Election):} \textit{Support Republican}, \textit{Support Democrat}, or \textit{Unclear}.
    \item \textbf{DEI (Social Issue):} \textit{Support Abolition of DEI}, \textit{Oppose Abolition of DEI}, or \textit{Unclear}.
    \item \textbf{FRIR (Interest Rate):} \textit{Support Interest Rate Cuts}, \textit{Oppose Interest Rate Cuts}, or \textit{Unclear}.
\end{itemize}
\subsection{Recruitment and Quality Control}
To ensure the depth and accuracy of cognitive labeling, we recruited 25 graduate students with backgrounds in Psychology and Cognitive Science (21 Master's candidates and 4 Ph.D. candidates). The annotation process spanned 84 days.

We implemented a rigorous training and quality control pipeline:
\begin{enumerate}
    \item \textbf{Pilot Training:} Before the main phase, annotators underwent training with 50 "Gold Standard" samples. We iteratively refined the guidelines and conducted calibration sessions until the inter-annotator agreement (Cohen's Kappa) exceeded \textbf{0.75}.
    \item \textbf{Consensus Mechanism:} Each sample in the dataset was independently labeled by three annotators. We adopted a majority vote strategy: a label was accepted if at least 2 out of 3 annotators agreed.
    \item \textbf{Strict Filtering:} To maintain high reliability, any sample where all three annotators disagreed (0/3 agreement) on any single dimension was discarded. 
\end{enumerate}
\subsection{Dataset Distribution Analysis}

\label{app:distribution}

Figure~\ref{fig:appendix_pie_charts} (see Appendix Figures) visualizes the comprehensive distribution of cognitive labels across the three domains. Below, we provide a detailed analysis of the cognitive characteristics observed in each dataset.

\paragraph{CUT (China-US Trade).}
The discourse in the CUT domain is characterized by a relatively detached and observational tone. In terms of \textit{Emotion}, \textit{Neutral} is the dominant category (48.72\%), significantly higher than \textit{Disgust} (22.01\%) and \textit{Anger} (12.04\%), suggesting that users often approach trade policy discussions with less immediate affective intensity. This rational tendency is mirrored in the \textit{Thinking} dimension: while \textit{Subjective Evaluation} (32.87\%) remains the most common value, the dataset exhibits a notable presence of \textit{Logical} (16.33\%) and \textit{Evidence-based} (11.14\%) reasoning compared to other topics. Regarding \textit{Intent}, \textit{Opinion Expression} accounts for 51.70\%, with a relatively low rate of \textit{Disagreement \& Conflict} (13.77\%). The \textit{Stance} distribution reveals a high degree of ambiguity, with 56.33\% of posts labeled as \textit{Unclear}, while explicit support leans towards China (33.49\%) over the US (10.17\%).

\paragraph{UE (US Election).}
In contrast, the UE dataset exhibits strong affective polarization and intuitive cognition. The \textit{Emotion} dimension is led by \textit{Disgust} (39.62\%), followed by \textit{Neutral} (34.40\%) and \textit{Anger} (10.10\%), reflecting the intense aversion often found in partisan politics. Cognitively, users rely heavily on System 1 thinking, with \textit{Subjective Evaluation} (41.00\%) and \textit{Emotional Judgment} (21.72\%) constituting the majority of thinking values. The \textit{Intent} distribution shows a significant rise in adversarial interaction, with \textit{Disagreement \& Conflict} reaching 25.45\%, although \textit{Opinion Expression} (48.28\%) remains the primary communicative goal. In terms of \textit{Stance}, the dataset shows a clear partisan divide, with 40.62\% supporting the Republican Party, 22.01\% supporting the Democratic Party, and 37.37\% remaining Unclear.

\paragraph{DEI (Diversity, Equity, and Inclusion).}
The DEI dataset represents the most intense conflict zone among the three topics. While \textit{Disgust} (42.22\%) reaches its peak here—surpassing both CUT and UE—\textit{Neutral} emotion is also substantial (36.76\%). The \textit{Thinking} patterns are diverse: \textit{Subjective Evaluation} (31.87\%) is lower than in the election topic, with users frequently employing \textit{Experience-based} (14.91\%) and \textit{Critical} (14.43\%) thinking, likely due to the personal nature of social identity debates. Notably, this topic has the highest density of \textit{Disagreement \& Conflict} (32.84\%) in the \textit{Intent} dimension, confirming its highly controversial nature. The \textit{Stance} distribution is heavily skewed towards the non-committed, with a massive 63.81\% labeled as \textit{Unclear}, while the remaining users are evenly split between opposing the abolition of DEI (19.15\%) and supporting it (17.03\%).

\paragraph{FRIR (Federal Reserve Interest Rate).}
The FRIR dataset exhibits a unique blend of economic rationality and livelihood anxiety. 
In terms of \textit{Stance}, it is the most balanced domain, with users almost equally divided between \textit{Support Interest Rate Cuts} (31.34\%) and \textit{Oppose Interest Rate Cuts} (30.14\%), reflecting the genuine complexity of the economic policy debate.
The \textit{Emotion} distribution is dominated by \textit{Neutral} (39.21\%) and \textit{Disgust} (31.74\%), suggesting a tone that oscillates between objective analysis and dissatisfaction with the economic status quo.
Cognitively, while \textit{Subjective Evaluation} (48.69\%) is prevalent (likely driven by personal financial stress), there is a significant presence of analytical thinking, including \textit{Logical} (14.50\%) and \textit{Evidence-based} (10.45\%) reasoning.
Regarding \textit{Intent}, the discourse is primarily focused on \textit{Opinion Expression} (57.31\%), with a moderate level of \textit{Disagreement} (13.24\%), indicating a constructive yet divided discussion environment.

\begin{figure*}[t]
    \centering
    \includegraphics[width=0.95\linewidth]{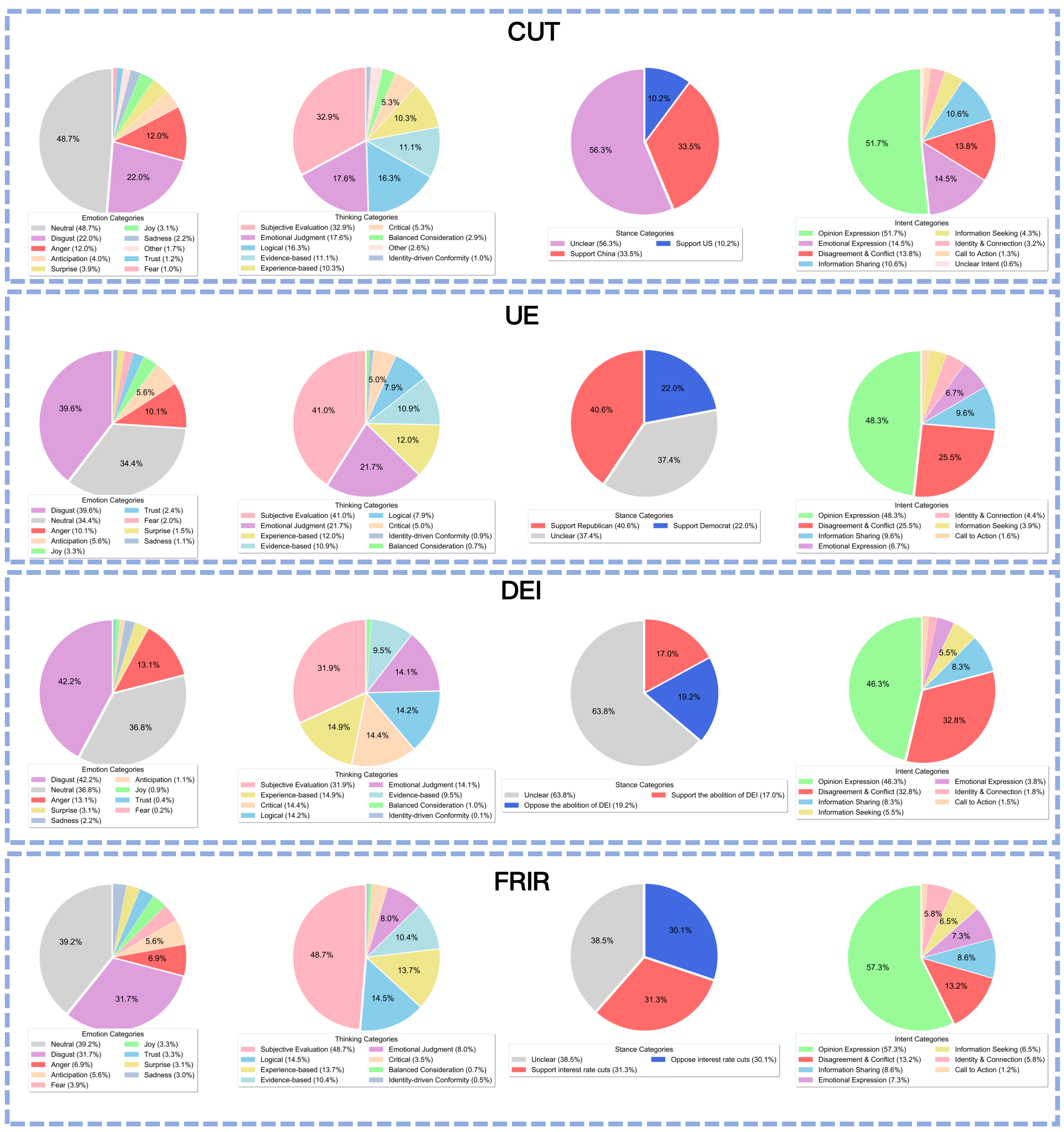} 
    \caption{Distribution of labels across Emotion, Thinking, Stance, and Intent dimensions for the four datasets (CUT, UE, DEI, and FRIR).}
    \label{fig:appendix_pie_charts}
\end{figure*}







\subsection{Intrinsic Hyperbolicity of Cognitive States}
\label{sec:gromov_analysis}

Before detailing our proposed method, we verify a fundamental hypothesis: \textit{Do complex cognitive states naturally reside in a hyperbolic manifold?} To answer this, we performed a topological analysis on our CognitiveBench dataset across its four distinct domains: CUT (China-U.S. Trade), DEI (Diversity, Equity, and Inclusion), UE (U.S. Election), and FRIR (Federal Reserve Interest Rates).

We utilized the Gromov $\delta$-hyperbolicity metric to quantify the "tree-likeness" of the semantic spaces formed by these discussions. In geometric group theory, a metric space is $\delta$-hyperbolic if, for any geodesic triangle, each side is contained within the $\delta$-neighborhood of the other two sides. A value of $\delta \to 0$ indicates that the underlying data structure is mathematically isomorphic to a tree.


As shown in Figure~\ref{fig:gromov_analysis}, the results are compelling. Across all four domains, the mean Gromov $\delta$ remains consistently low (Mean $\delta \approx 0.015$). More importantly, the Relative Delta (defined as $2\delta / \text{diam}(\mathcal{X})$, where $\text{diam}(\mathcal{X})$ is the maximum geodesic distance between any pair of points in the dataset) is approximately $1.0\% - 1.1\%$ for all datasets (e.g., UE: $0.0101$, DEI: $0.0107$). 

These low relative $\delta$ values provide strong empirical evidence that human cognitive states—characterized by hierarchical intentions, stances, and emotions—exhibit high hyperbolicity. When such tree-like data is embedded into the flat Euclidean space of standard LLMs, it inevitably suffers from significant distortion. This finding serves as the cornerstone of our HyCoLLM framework, empirically validating the necessity of a hyperbolic geometric prior.

\section{Global Notation System}
\label{app:notation}

To ensure consistency throughout the paper, we summarize the core mathematical notations and their corresponding definitions in Table \ref{tab:notation}.

\begin{table*}[tb]
    \centering
    
    \renewcommand{\arraystretch}{1.1} 
    \resizebox{1.0\textwidth}{!}{ 
    \begin{tabular}{l l l l}
        \toprule
        \textbf{Concept} & \textbf{Symbol} & \textbf{Description} & \textbf{Space Property} \\
        \midrule
        Raw Input & $X = (C, T)$ & Context $C$ + Target Post $T$ & Discrete Text \\
        Hyperbolic Manifold & $\mathbb{D}_c^d$ & Poincaré ball of dim $d$ with curvature $-c$ & Hyperbolic Space \\
        Tangent Space (Origin) & $T_{\mathbf{0}}\mathbb{D}_c^d$ & Tangent space at origin (local approximation) & Euclidean Space $\mathbb{R}^d$ \\
        Exponential Map & $\exp_{\mathbf{0}}^c(\cdot)$ & Map from Tangent Space $\to$ Manifold & $\mathbb{R}^d \to \mathbb{D}_c^d$ \\
        Logarithmic Map & $\log_{\mathbf{0}}^c(\cdot)$ & Map from Manifold $\to$ Tangent Space & $\mathbb{D}_c^d \to \mathbb{R}^d$ \\
        HCN Intermediate & $\mathbf{z}_k$ & Feature for dimension $k$ (e.g., emotion) & $\in \mathbb{D}_c^d$ (On Manifold) \\
        HCN Final Output & $\mathbf{v}_{cog}$ & Cognitive Anchor & $\in T_{\mathbf{0}}\mathbb{D}_c^d \cong \mathbb{R}^d$ (Tangent) \\
        LLM Hidden State & $\mathbf{h}_{last}$ & Last layer hidden state of LLM & $\in \mathbb{R}^{d_{model}}$ (Euclidean) \\
        Semantic Projection & $\hat{\mathbf{v}}_{sem}$ & Projected vector for alignment & $\in \mathbb{R}^d$ (Tangent Space) \\

        Soft Prompt Sequence & $\mathbf{E}_{prior}$ & Prompt Embeddings generated by Projector & $\in \mathbb{R}^{L \times d_{model}}$ \\
        \bottomrule
    \end{tabular}}
\caption{Global Notation System and Geometric Properties.}
\label{tab:notation}
\end{table*}

\subsection{The Basic Formula of the Poincaré Sphere}
\label{app:basic_formula}

To capture the hierarchical structure of cognitive concepts (e.g., the intrinsic hierarchy from broad sentiments to specific emotions), we utilize the Poincaré ball model $(\mathbb{D}_c^d, g^{\mathbb{D}})$. The manifold is defined as $\mathbb{D}_c^d = \{\boldsymbol{z} \in \mathbb{R}^d : \|\boldsymbol{z}\| < 1/\sqrt{c}\}$, where $c$ is the curvature (set to $c=1$ in implementation). We map points between the Euclidean space $\boldsymbol{x} \in \mathbb{R}^d$ and the hyperbolic space $\boldsymbol{z} \in \mathbb{D}_c^d$ using the \textit{exponential map} ($\exp_{\mathbf{0}}^c$) and \textit{logarithmic map} ($\log_{\mathbf{0}}^c$) defined at the origin $\mathbf{0}$:

\begin{equation}
    \boldsymbol{z} = \exp_{\mathbf{0}}^c(\boldsymbol{x}) = \tanh(\sqrt{c}\|\boldsymbol{x}\|) \frac{\boldsymbol{x}}{\sqrt{c}\|\boldsymbol{x}\|}
\end{equation}

\begin{equation}
    \boldsymbol{x} = \log_{\mathbf{0}}^c(\boldsymbol{z}) = \frac{1}{\sqrt{c}}\operatorname{arctanh}(\sqrt{c}\|\boldsymbol{z}\|) \frac{\boldsymbol{z}}{\|\boldsymbol{z}\|}
\end{equation}

The induced distance between two points $\boldsymbol{z}_i, \boldsymbol{z}_j$ in the hyperbolic space (where $\|\boldsymbol{z}\| < 1/\sqrt{c}$) is defined as the Poincaré distance:

\begin{equation}
\small
d_{\mathbb{D}}(\boldsymbol{z}_i, \boldsymbol{z}_j) = \frac{1}{\sqrt{c}}\operatorname{arccosh}\left(1 + 2c \frac{\|\boldsymbol{z}_i - \boldsymbol{z}_j\|^2}{(1 - c\|\boldsymbol{z}_i\|^2)(1 - c\|\boldsymbol{z}_j\|^2)}\right)
\end{equation}

\section{Theoretical Analysis: Geometric Capacity Mismatch}
\label{app:theory}

In this section, we provide a rigorous theoretical analysis supporting Proposition~1 in Section~\ref{sec:theory_overview}. We formalize the geometric capacity mismatch between Euclidean and hyperbolic spaces when embedding hierarchical cognitive structures.

\subsection{Preliminaries}

\textbf{Definition A.1 (Hierarchical Cognitive Taxonomy).}
\textit{A hierarchical cognitive taxonomy $\mathcal{T}$ is modeled as a rooted tree with branching factor $b>1$ and depth $k$. The number of nodes at depth $k$ is $|\mathcal{T}_k| = b^k$.}

\textbf{Definition A.2 (Bounded-Distortion Embedding).}
\textit{An embedding $f : \mathcal{T} \rightarrow (\mathcal{M}, d)$ into a metric space $(\mathcal{M}, d)$ is said to have bounded distortion if there exists $\epsilon>0$ such that the pairwise distance between any two distinct nodes at the same depth is at least $\epsilon$.}

\textbf{Assumption A.1 (Finite Dimensionality).}
\textit{We consider embeddings into $d$-dimensional spaces, where $d$ is fixed and finite.}
\subsection{Euclidean Embedding Capacity}

\textbf{Lemma A.1 (Polynomial Volume Growth of Euclidean Space).}
\textit{Let $B_{\mathbb{E}}(R) \subset \mathbb{R}^d$ be a Euclidean ball of radius $R$. Its volume satisfies}
\begin{equation}
    \mathrm{Vol}_{\mathbb{E}}(R) = C_d R^d ,
\end{equation}
\textit{where $C_d>0$ is a constant depending only on $d$.}

\textbf{Theorem A.1 (Exponential Radius Requirement in Euclidean Space).}
\textit{Let $\mathcal{T}$ be a hierarchical taxonomy with branching factor $b>1$ and depth $k$. Any bounded-distortion embedding of $\mathcal{T}$ into $\mathbb{R}^d$ requires the embedding radius $R$ to grow exponentially with $k$, i.e.,}
\begin{equation}
    R = \Omega\!\left(b^{k/d}\right).
\end{equation}

\begin{proof}
At depth $k$, the taxonomy contains $b^k$ nodes. By Definition~A.2, these nodes must be embedded within a ball of radius $R$ such that each pair of nodes is separated by at least $\epsilon$.

Packing arguments imply that the maximum number of $\epsilon$-separated points in a Euclidean ball of radius $R$ is upper bounded by $O(R^d)$. Therefore, we require:
\begin{equation}
    R^d \ge C \cdot b^k ,
\end{equation}
for some constant $C>0$. Taking the $d$-th root yields:
\begin{equation}
    R \ge C^{1/d} b^{k/d}.
\end{equation}
Thus, the embedding radius must grow exponentially with the hierarchy depth $k$, completing the proof.
\end{proof}

\subsection{Hyperbolic Embedding Capacity}

\textbf{Definition A.3 (Poincaré Ball Model).}
\textit{Let $\mathbb{D}_c^d = \{ \boldsymbol{x} \in \mathbb{R}^d : \|\boldsymbol{x}\| < 1/\sqrt{c} \}$ denote the $d$-dimensional Poincaré ball with constant negative curvature $-c$, equipped with the standard hyperbolic metric.}

\textbf{Lemma A.2 (Exponential Volume Growth of Hyperbolic Space).}
\textit{Let $B_{\mathbb{H}}(R) \subset \mathbb{D}_c^d$ be a hyperbolic ball of radius $R$. Its volume satisfies}
\begin{equation}
    \mathrm{Vol}_{\mathbb{H}}(R)
    = \Omega_{d-1} \int_0^R \sinh^{d-1}(\sqrt{c} r)\,dr ,
\end{equation}
\textit{and for sufficiently large $R$,}
\begin{equation}
    \mathrm{Vol}_{\mathbb{H}}(R) = \Theta\!\left(e^{(d-1)\sqrt{c}R}\right).
\end{equation}

\textbf{Theorem A.2 (Linear Radius Requirement in Hyperbolic Space).}
\textit{Let $\mathcal{T}$ be a hierarchical taxonomy with branching factor $b>1$ and depth $k$. There exists a bounded-distortion embedding of $\mathcal{T}$ into $\mathbb{D}_c^d$ whose embedding radius grows linearly with $k$, i.e.,}
\begin{equation}
    R = O(k).
\end{equation}

\begin{proof}
At depth $k$, the taxonomy contains $b^k$ nodes. By Lemma~A.2, the volume of a hyperbolic ball of radius $R$ grows exponentially with $R$. To accommodate $b^k$ $\epsilon$-separated nodes, it suffices that:
\begin{equation}
    e^{(d-1)\sqrt{c}R} \ge C' \cdot b^k ,
\end{equation}
for some constant $C'>0$. Taking logarithms yields:
\begin{equation}
    R \ge \frac{k \log b + \log C'}{(d-1)\sqrt{c}} .
\end{equation}
Thus, the required embedding radius grows linearly with the depth $k$, completing the proof.
\end{proof}

\subsection{Implications for Cognitive Crowding}

\textbf{Corollary A.1 (Geometric Origin of Cognitive Crowding).}
\textit{Under bounded-distortion constraints, Euclidean space cannot embed deep hierarchical cognitive taxonomies without exponential growth in radius, leading to inevitable crowding of semantically distinct states. Hyperbolic space avoids this limitation due to its exponential volume growth.}

This corollary establishes a necessary geometric condition underlying cognitive crowding and provides formal justification for adopting hyperbolic representations in HyCoLLM.

\section{Supplementary Experiment}
\label{app:suppl_exp}
\subsection{Detailed Experimental Setup}
\label{app:exp_details}

\subsubsection{Hyperbolic Cognitive Network (HCN) Configuration}
The HCN serves as the geometric teacher and is composed of three key modules:
\begin{itemize}
\item \textbf{Hyperbolic Projection:} We use a Poincaré ball manifold with curvature $c=1.0$. Input features (dim=4096) are projected to a hidden dimension of 512 using an exponential map.
\item \textbf{Cognitive Transformer:} The encoder consists of $N=4$ layers with $H=8$ attention heads and a dropout rate of 0.1. A learned dimension embedding is added to distinguish between the four cognitive states before cross-dimension attention.
\item \textbf{Multi-Task Heads:} We utilize separate Multi-Layer Perceptrons (MLPs) for each dimension: Emotion (9 classes), Thinking (8 classes), Intent (7 classes), and Stance (3 classes).
\end{itemize}
The HCN is pre-trained using a Riemannian Adam optimizer with a learning rate of $1e-3$ and a batch size of 32. The loss function combines task-specific cross-entropy losses with a hyperbolic geometric regularization ($\lambda_{hyper}=0.1$) and contrastive loss ($\lambda_{contrastive}=0.1$) with a temperature of $\tau=0.5$.

\subsubsection{Hyperbolic-Guided Alignment Tuning (HGAT)}
For the alignment phase, we freeze the HCN and tune the LLM (Llama3.1-8b/Qwen3-8b).
\begin{itemize}
\item \textbf{Projectors:} The \textit{Cognitive Projector} maps the 512-dim HCN anchor to the LLM's embedding space via a 2-layer MLP with Tanh activation to bound the soft prompt values. The \textit{Alignment Projector} maps the LLM's last hidden state back to the 512-dim tangent space.
\item \textbf{Training Hyperparameters:} We employ 4-bit NormalFloat (NF4) quantization and QLoRA with rank $r=16$, $\alpha=32$, and dropout 0.0. The model is trained for 3 epochs with a learning rate of $2e-4$, a batch size of 1, and gradient accumulation steps of 4. The SCT loss weight is set to $\lambda=1.0$ based on sensitivity analysis.
\item \textbf{Environment:} All experiments were conducted on a single NVIDIA H800 (80GB) GPU using PyTorch 2.1 and the \texttt{vllm} library for inference acceleration.
\end{itemize}

\subsection{Computational Efficiency}
\label{app:efficiency}
\begin{figure}[t]
    \centering
    \includegraphics[width=\linewidth]{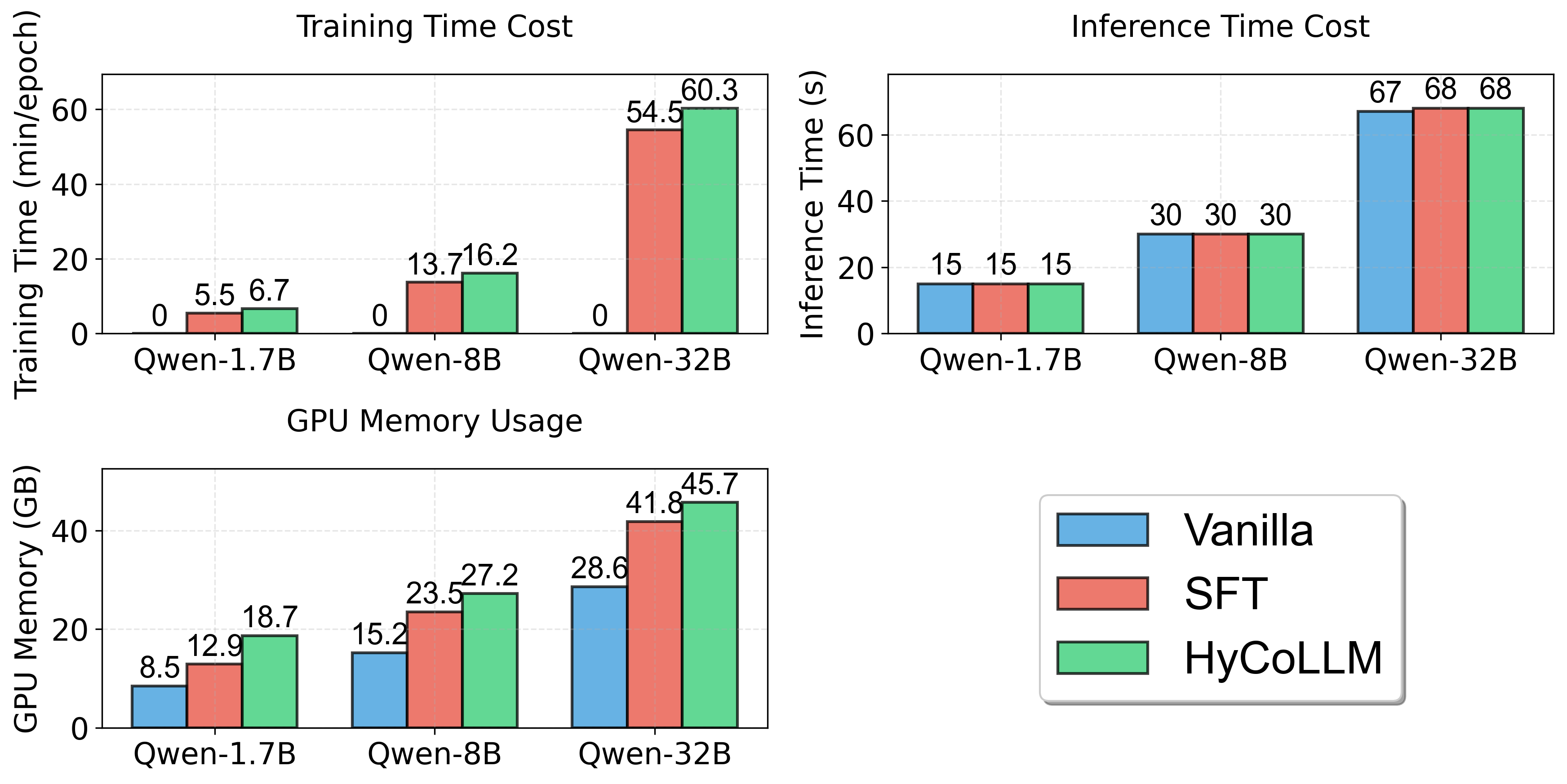}
    \caption{Computational Cost Analysis across different model sizes. While HyCoLLM introduces a marginal increase in training time and memory usage due to the additional optimization of geometric priors, it incurs zero additional latency during inference compared to SFT, ensuring deployment efficiency.}
    \label{fig:cost_analysis}
\end{figure}


As illustrated in Figure \ref{fig:cost_analysis}, HyCoLLM introduces a moderate training overhead due to the dual-optimization process, with training time increasing by approximately 10\%$\sim$18\%. Crucially, this overhead exhibits favorable scalability: as the model size increases from 1.7B to 32B, the relative time cost diminishes (from $\sim$21\% to $\sim$10\%), and the additional GPU memory usage remains roughly constant ($\approx 3.7 \sim 4.0$ GB) rather than growing linearly. This indicates that the geometric component acts as a fixed-cost module, making the method increasingly cost-effective for larger foundation models.

Regarding deployment, HyCoLLM achieves \textbf{zero additional latency} compared to vanilla SFT models (e.g., maintaining constant 30s inference time for Qwen3-8B). This is because the complex Riemannian computations are strictly confined to the training phase to construct the geometric prior. During inference, the HCN module is detached, and the structural constraints are implicitly preserved within the optimized soft prompts and parameters. Consequently, the model generates topologically consistent responses without requiring specialized Riemannian kernels or extra floating-point operations at runtime.

\subsection{Hyperparameter Study}
\label{app:hyperparams}
\begin{figure}[t]
    \centering
    \includegraphics[width=0.9\linewidth]{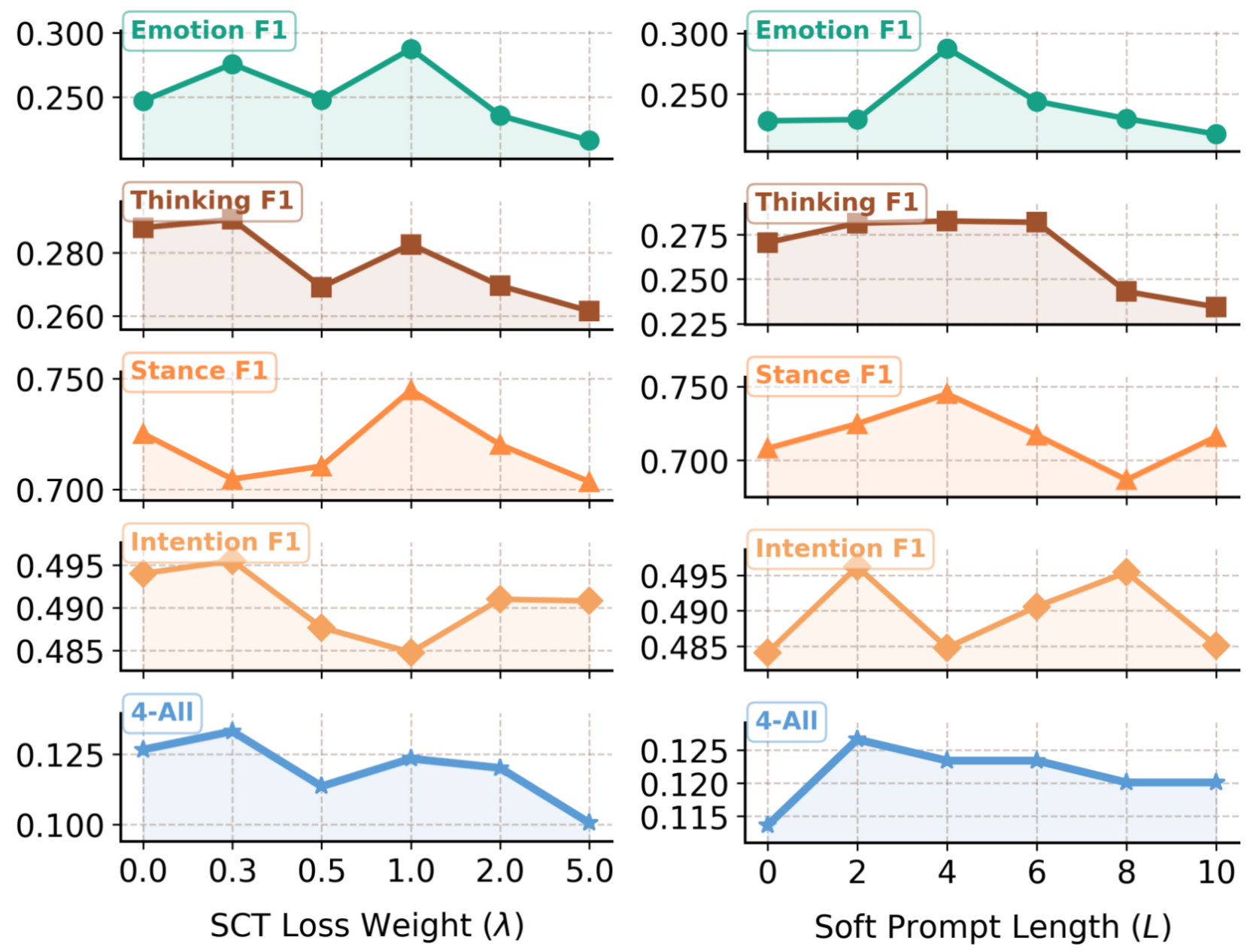} 
    \caption{Hyperparameter sensitivity analysis averaged across three datasets. The plots illustrate the impact of the SCT loss weight $\lambda$ (Left) and the soft prompt length $L$ (Right) on dimension-specific F1 scores and the holistic PMA@4 accuracy. Both parameters exhibit a distinct inverted U-shape trend, demonstrating that $\lambda=1.0$ and $L=4$ provide the optimal balance for HyCoLLM.}
\label{fig:hyperparams}
\end{figure}

We investigate the sensitivity of the Semantic-Cognitive Topology loss weight ($\lambda$) and soft prompt length ($L$) in Figure \ref{fig:hyperparams}, where both metrics exhibit a distinct inverted U-shape trend. Regarding the loss weight, performance peaks at $\lambda=1.0$, suggesting that geometric structural constraints and auto-regressive generation are of equal importance; lower values ($\lambda<0.5$) fail to sufficiently correct Euclidean crowding, while higher values ($\lambda>2.0$) over-regularize the semantic space, a stable balance further corroborated by the convergence curves in Appendix \ref{app:training_dynamics}. Simultaneously, for the prompt length, we observe that performance improves as $L$ increases to 4 but degrades thereafter. This indicates that $L=4$ offers the optimal trade-off, providing sufficient token capacity to encode complex hyperbolic coordinates while avoiding the feature dilution and noise interference associated with excessively long prompts.

\subsection{Intuitive Explanation of Hyperbolic Space}
\label{app:intuition}

To make the geometric intuition more accessible, we elaborate on the ``tree-on-a-saddle'' analogy introduced in the main text. Consider a deeply branching cognitive taxonomy $\mathcal{T}$: emotions split into positive, negative, and neutral categories, each of which further branches into fine-grained labels such as Anger, Disgust, Fear, and Sadness; similar branching occurs for thinking styles, stances, and intents. The number of leaves grows exponentially with depth ($b^k$), while the ``volume'' available to separate these leaves in a $d$-dimensional Euclidean ball grows only polynomially ($R^d$). Intuitively, as we try to embed this tree into flat space, deeper nodes are forced into overlapping regions: semantically distinct cognitive states collapse onto each other, yielding \textit{Cognitive Crowding}.

Hyperbolic space behaves very differently. The surface area of a hyperbolic ball grows \textit{exponentially} with radius, mirroring the exponential growth of the taxonomy. A helpful visualization is the ruffled edge of a kale leaf or a saddle surface: as one moves outward from the center, there is always ``more room'' for new branches. When we place the cognitive taxonomy on such a surface, every leaf can be assigned a distinct region even for deep hierarchies. HyCoLLM's Hyperbolic Cognitive Network plays the role of ``embedding the tree on the saddle'', while the Hyperbolic-Guided Alignment Tuning aligns the LLM's Euclidean representations with this more spacious geometry. The formal version of this intuition is provided in Section~\ref{sec:theory_overview} and Appendix~\ref{app:theory}.

\subsection{Vanilla and Few-shot CoT Baselines}
\label{app:cot_baselines}

To assess whether prompting strategies can close the gap to HyCoLLM without parameter updates, we evaluate Qwen3-8B and Llama3.1-8B under two additional settings: \textbf{Vanilla} (zero-shot) and \textbf{Vanilla+CoT} (5-shot chain-of-thought), both using the same 4-dimensional prompt template. Table~\ref{tab:cot_baselines} summarizes the results.

\begin{table*}[t]
\centering

\resizebox{\textwidth}{!}{%
\begin{tabular}{ll | cc cc cc cc | ccc | c}
\toprule
\multirow{2}{*}{Backbone} & \multirow{2}{*}{Setting} & \multicolumn{2}{c}{Emotion} & \multicolumn{2}{c}{Thinking} & \multicolumn{2}{c}{Stance} & \multicolumn{2}{c|}{Intent} & \multirow{2}{*}{PMA@2} & \multirow{2}{*}{PMA@3} & \multirow{2}{*}{PMA@4} & \multirow{2}{*}{H-Loss $\downarrow$} \\
 & & ACC & F1 & ACC & F1 & ACC & F1 & ACC & F1 & & & & \\
\midrule
\multirow{4}{*}{Qwen3-8B}
 & Vanilla       & 0.4123 & 0.1608 & 0.2630 & 0.1752 & 0.2370 & 0.2076 & 0.4448 & 0.4342 & 0.4286 & 0.1169 & 0.0097 & 0.1803 \\
 & Vanilla+CoT   & 0.5617 & 0.1914 & 0.2922 & 0.1842 & 0.6623 & 0.6678 & 0.5130 & 0.4664 & 0.7110 & 0.3214 & 0.0455 & 0.1450 \\
 & LoRA-SFT      & 0.5844 & 0.2525 & 0.4383 & 0.2484 & 0.6721 & 0.6582 & 0.5812 & 0.4294 & 0.7500 & 0.4286 & 0.1396 & 0.1272 \\
 & \textbf{HyCoLLM} & \textbf{0.6266} & 0.2485 & \textbf{0.4513} & \textbf{0.2813} & \textbf{0.7338} & \textbf{0.7425} & \textbf{0.5877} & \textbf{0.4640} & \textbf{0.8052} & \textbf{0.4903} & \textbf{0.1396} & \textbf{0.1183} \\
\midrule
\multirow{4}{*}{Llama3.1-8B}
 & Vanilla       & 0.1466 & 0.0521 & 0.2378 & 0.0913 & 0.5081 & 0.5009 & 0.5000 & 0.1717 & 0.4365 & 0.1285 & 0.0098 & 0.1927 \\
 & Vanilla+CoT   & 0.1851 & 0.1078 & 0.1623 & 0.1526 & 0.4513 & 0.3959 & 0.4838 & 0.2526 & 0.3669 & 0.1039 & 0.0097 & 0.1778 \\
 & LoRA-SFT      & 0.5270 & 0.3256 & 0.3703 & 0.1763 & 0.6392 & 0.5543 & 0.5366 & 0.3351 & 0.7047 & 0.3480 & 0.0795 & 0.1516 \\
 & \textbf{HyCoLLM} & \textbf{0.6133} & \textbf{0.3289} & \textbf{0.5091} & \textbf{0.3771} & \textbf{0.6508} & 0.5543 & \textbf{0.6584} & \textbf{0.5392} & \textbf{0.8179} & \textbf{0.5047} & \textbf{0.1431} & \textbf{0.1215} \\
\bottomrule
\end{tabular}}
\caption{Vanilla and Vanilla+CoT baselines compared with LoRA-SFT and HyCoLLM on the CognitiveBench test set. ACC and F1 are reported for each dimension; PMA@\textit{k} and Hamming Loss measure holistic consistency. Lower is better ($\downarrow$) for Hamming Loss.}
\label{tab:cot_baselines}
\end{table*}

Two observations stand out. First, Vanilla+CoT yields only marginal gains on Qwen3-8B and is actually \textit{detrimental} for Llama3.1-8B, where Stance ACC drops from 0.5081 (Vanilla) to 0.4513 (Vanilla+CoT). Error analysis shows that without parameter tuning, general-purpose 8B models frequently fail to respect the strict JSON schema required for the 4-dimensional taxonomy, leading to parsing errors or hallucinated labels. Second, LoRA-SFT and especially HyCoLLM dominate both Vanilla and CoT variants on every holistic metric, with HyCoLLM reaching 13.96\%/14.31\% PMA@4 on Qwen3-8B/Llama3.1-8B versus at most 4.55\%/0.97\% for CoT prompting. This validates the necessity of geometric alignment over purely in-context strategies for strict multi-dimensional cognitive modeling.

\subsection{Two-Path Cumulative Ablation}
\label{app:cumulative_ablation}

The ablation in Table~\ref{tab:ablation_avg} removes a single component at a time relative to the full model. To further disentangle how the Soft Prompt and the Semantic-Cognitive Topology (SCT) loss interact, we conduct a rigorous \textit{two-path cumulative ablation}: starting from the full HyCoLLM-Qwen3-8B, we progressively remove one component and then the other, in two different orders.

\begin{table}[t]
\centering
\small

\begin{tabular}{clcc}
\toprule
\textbf{Step} & \textbf{Model state} & \textbf{PMA@4} & \textbf{$\Delta$} \\
\midrule
\multicolumn{4}{l}{\emph{Path A: Remove Soft Prompt $\rightarrow$ Remove SCT}} \\
0 & Full HyCoLLM           & 15.49 & -- \\
1 & w/o Soft Prompt (only SCT) & 11.76 & $-3.73$ \\
2 & w/o Soft Prompt \& SCT (SFT) & 10.91 & $-0.85$ \\
\midrule
\multicolumn{4}{l}{\emph{Path B: Remove SCT $\rightarrow$ Remove Soft Prompt}} \\
0 & Full HyCoLLM           & 15.49 & -- \\
1 & w/o SCT (only Soft Prompt) & 12.52 & $-2.97$ \\
2 & w/o Soft Prompt \& SCT (SFT) & 10.91 & $-1.61$ \\
\bottomrule
\end{tabular}
\caption{Two-path cumulative ablation of the Soft Prompt and SCT loss, reported in PMA@4 (\%). Both paths converge to the LoRA-SFT baseline when both components are removed.}
\label{tab:cumulative_ablation}
\end{table}

Table~\ref{tab:cumulative_ablation} reveals two effects. (i) \textbf{Dominance of the Soft Prompt.} Comparing the intermediate states, the model with \textit{only} the Soft Prompt (12.52\%) outperforms the one with \textit{only} the SCT loss (11.76\%), indicating that injecting the hyperbolic geometric prior at the input stage is more effective than the output-side alignment alone. (ii) \textbf{Synergistic effect.} Removing either component causes a substantial drop from the full model (15.49\%), and both paths converge to the same SFT baseline (10.91\%) when both are removed. The drops are non-additive: the two modules are complementary, with the Soft Prompt providing the geometric context and the SCT loss enforcing structural consistency of the predictions.

\subsection{Parameter Scalability Analysis}
\label{app:scaling}

We investigate the scalability of our approach by applying HyCoLLM to the Qwen3 model family with varying parameter sizes: 1.7B, 8B, and 32B. We compare three settings: Vanilla (Base Model), SFT, and HyCoLLM. Table \ref{tab:scaling_full} details the performance across all metrics.

\paragraph{Analysis of Model Scale.}
As shown in Table \ref{tab:scaling_full}, HyCoLLM consistently outperforms the SFT baseline across all parameter scales, validating the universality of the geometric prior.
\begin{itemize}
    \item \textbf{Small Models (1.7B):} The improvement is most substantial here. HyCoLLM nearly doubles the \textit{PMA@4} accuracy of SFT (from 0.0584 to 0.1071) and significantly reduces \textit{Hamming Loss}. This supports our hypothesis that hyperbolic geometry effectively "expands" the embedding capacity of smaller models, allowing them to disentangle complex cognitive states that usually require larger parameter spaces.
    \item \textbf{Large Models (32B):} Even with the powerful Qwen3-32B, HyCoLLM maintains a clear edge over SFT (\textit{Hamming Loss} 0.1255 vs. 0.1340). While the base capabilities of 32B are strong, SFT still struggles with cognitive crowding (PMA@4 = 0.0942). HyCoLLM rectifies this, proving that geometric structural constraints provide value beyond simple parameter scaling.
\end{itemize}

\begin{table*}[h]
\centering

\resizebox{\textwidth}{!}{
\begin{tabular}{cc|cc|cc|cc|cc|ccc|c}
\toprule
\multirow{2}{*}{Model} & \multirow{2}{*}{Setting} & \multicolumn{2}{c|}{Emotion} & \multicolumn{2}{c|}{Thinking} & \multicolumn{2}{c|}{Stance} & \multicolumn{2}{c|}{Intent} & \multirow{2}{*}{PMA@2} & \multirow{2}{*}{PMA@3} & \multirow{2}{*}{PMA@4} & \multirow{2}{*}{H-Loss $\downarrow$} \\
 & & ACC & F1 & ACC & F1 & ACC & F1 & ACC & F1 & & & & \\ 
\midrule
\multirow{3}{*}{Qwen3-1.7B} 
 & Vanilla & 0.2825 & 0.0854 & 0.2435 & 0.1348 & 0.4870 & 0.4301 & 0.3019 & 0.1741 & 0.4188 & 0.1331 & 0.0097 & 0.1984 \\
 & SFT & 0.4221 & 0.1741 & 0.3539 & 0.1077 & 0.5812 & 0.5430 & 0.5065 & 0.2111 & 0.6169 & 0.2857 & 0.0584 & 0.1576 \\
 & \textbf{HyCoLLM} & \textbf{0.5552} & \textbf{0.1959} & \textbf{0.4513} & \textbf{0.2040} & \textbf{0.6461} & \textbf{0.6422} & \textbf{0.5487} & \textbf{0.3270} & \textbf{0.7565} & \textbf{0.3961} & \textbf{0.1071} & \textbf{0.1329} \\
\midrule
\multirow{3}{*}{Qwen3-8B} 
 & Vanilla & 0.4123 & 0.1608 & 0.2630 & 0.1752 & 0.2370 & 0.2076 & 0.4448 & 0.4342 & 0.4286 & 0.1169 & 0.0097 & 0.1803 \\
 & SFT & 0.5844 & \textbf{0.2525} & 0.4383 & 0.2484 & 0.6721 & 0.6582 & 0.5812 & 0.4294 & 0.7500 & 0.4286 & 0.1396 & 0.1272 \\
 & \textbf{HyCoLLM} & \textbf{0.6266} & 0.2485 & \textbf{0.4513} & \textbf{0.2813} & \textbf{0.7338} & \textbf{0.7425} & \textbf{0.5877} & \textbf{0.4640} & \textbf{0.8052} & \textbf{0.4903} & \textbf{0.1396} & \textbf{0.1183} \\
\midrule
\multirow{3}{*}{Qwen3-32B} 
 & Vanilla & 0.4383 & 0.2803 & 0.2695 & 0.1741 & 0.5649 & 0.5699 & 0.5390 & 0.4078 & 0.6299 & 0.2662 & 0.0260 & 0.1617 \\
 & SFT & 0.5617 & 0.2559 & 0.3604 & 0.1977 & \textbf{0.7240} & \textbf{0.7380} & 0.5390 & 0.4367 & 0.7532 & 0.3766 & 0.0942 & 0.1340 \\
 & \textbf{HyCoLLM} & \textbf{0.5714} & \textbf{0.2893} & \textbf{0.4351} & \textbf{0.2312} & 0.7078 & 0.7127 & \textbf{0.5812} & \textbf{0.5002} & \textbf{0.7857} & \textbf{0.4448} & \textbf{0.1071} & \textbf{0.1255} \\
\bottomrule
\end{tabular}
}
\caption{Performance comparison across different model sizes (Qwen3 family) on the CognitiveBench dataset. \textbf{Bold} indicates the best performance within each model size group.}
\label{tab:scaling_full}
\end{table*}

\subsection{Results of the sub-dataset}
\label{app:sub_main_results}
To further verify the robustness of HyCoLLM across varying cognitive contexts, we analyze the performance on three distinct sub-datasets: CUT (Economic), UE (Politic), and DEI (Culture).

\paragraph{Performance on CUT (Table \ref{tab:main_results}).}
The CUT dataset, characterized by rational economic discourse, presents a unique challenge where "Stance" and "Thinking" are often subtly intertwined with objective data. 
As shown in Table \ref{tab:main_results}, standard SFT models (e.g., Qwen3-8B) perform competitively, achieving the highest \textit{PMA@4} accuracy (13.96\%). This suggests that in more structured and fact-based domains, Euclidean representations can capture rigid cognitive patterns to some extent. 
However, \textbf{HyCoLLM} still demonstrates superior overall reliability. Specifically, our method achieves the lowest \textit{Hamming Loss} (0.1165) and significantly outperforms all baselines in \textit{Stance Accuracy} (73.70\% vs. GPT-4o 66.88\%) and \textit{PMA@2} consistency (81.49\%). This indicates that while SFT may memorize specific patterns, HyCoLLM offers a more generalized understanding of the "Economic Stance," effectively decoupling the complex dependency between a user's analytical thinking style and their support for trade policies.

\paragraph{Performance on UE (Table \ref{tab:ue_results}).}
In the US Election, the "Cognitive Crowding" phenomenon becomes acute for baseline models.
Table \ref{tab:ue_results} reveals that while closed-source giants like GPT-4o achieve a respectable \textit{Stance} accuracy (69.01\%), their holistic cognitive understanding collapses, with a \textit{PMA@4} score of only 5.05\%. This implies they can classify \textit{who} a user supports but fail to understand \textit{why} (Thinking/Emotion) and \textit{what they want} (Intent).
In contrast, \textbf{HyCoLLM} thrives in this hierarchical cognitive environment. It improves the \textit{PMA@4} accuracy to 14.95\%—nearly tripling the performance of GPT-4o—and achieves state-of-the-art results in \textit{Emotion} (62.64\%) and \textit{Thinking} (53.41\%). This substantial gain verifies our hypothesis: political cognition, structured as "Ideology $\rightarrow$ Stance $\rightarrow$ Expression," is naturally isomorphic to hyperbolic geometry, allowing HyCoLLM to disentangle the deep-seated emotional logic from surface-level partisanship.

\paragraph{Performance on DEI (Table \ref{tab:dei_results}).}
The DEI dataset represents the most complex "value conflict" zone, involving abstract social concepts and intense identity debates.
Results in Table \ref{tab:dei_results} highlight the most dramatic superiority of our approach. Standard SFT models struggle significantly here, with Qwen3-8B's \textit{PMA@4} accuracy dropping to 7.77\%. Conversely, \textbf{HyCoLLM} maintains exceptional consistency, achieving a \textit{PMA@4} accuracy of 19.17\%, which is \textbf{2.5$\times$ higher} than the strongest open-source baseline. Furthermore, HyCoLLM dominates the abstract \textit{Thinking} dimension (Acc 58.55\% vs. SFT 39.38\%).
This breakthrough suggests that purely semantic-based learning (SFT) falters when dealing with high-dimensional social values. By injecting geometric priors, HyCoLLM effectively constructs a "Cognitive Manifold" for diversity and inclusion, enabling it to navigate the intricate web of moral judgments and identity assertions with human-level nuance.

\begin{table*}[htpb]
\centering

\resizebox{\textwidth}{!}{
\begin{tabular}{ll | cc cc cc cc | ccc | c }
\toprule
\multicolumn{2}{c|}{\multirow{2}{*}{Model}} & 
\multicolumn{2}{c}{Emotion} & 
\multicolumn{2}{c}{Thinking} & 
\multicolumn{2}{c}{Stance} & 
\multicolumn{2}{c|}{Intent} & 
\multirow{2}{*}{PMA@2} & 
\multirow{2}{*}{PMA@3} & 
\multirow{2}{*}{PMA@4} & 
\multirow{2}{*}{\textit{Hamming Loss} $\downarrow$} \\
\multicolumn{2}{c|}{} & ACC & F1 & ACC & F1 & ACC & F1 & ACC & F1 & & & & \\ 
\midrule

\rowcolor{header_blue} \multicolumn{14}{c}{\textbf{Human}} \\
\multicolumn{2}{c|}{Averaged from 3 experts} & 0.7619 & 0.5768 & 0.6524 & 0.5183 & 0.8235 & 0.6633 & 0.7437 & 0.6054 & 0.9223 & 0.7172 & 0.3527 & 0.0676 \\
\midrule
\rowcolor{header_blue} \multicolumn{14}{c}{\textbf{Close-Source}} \\
\multirow{4}{*}{\shortstack[l]{Small-size\\Few-Shot \& CoT}} 
 & GPT4o-mini & 0.4578 & 0.2082 & 0.2922 & 0.1615 & 0.6201 & 0.6297 & 0.5779 & 0.4262 & 0.6656 & 0.3117 & 0.0455 & 0.1504 \\
 & Gemini2.5-Flash-Lite & 0.5928 & 0.2928 & 0.2834 & 0.2085 & 0.6254 & 0.6411 & 0.5342 & 0.4408 & 0.7231 & 0.3225 & 0.0554 & 0.1453 \\
 & Grok-4-fast & 0.5010 & 0.2589 & 0.2908 & 0.2028 & 0.5541 & 0.5293 & 0.5180 & 0.4750 & 0.6503 & 0.2680 & 0.0196 & 0.1572 \\
 & Qwen-Turbo & 0.5714 & 0.1989 & 0.3019 & 0.2037 & 0.6331 & 0.6443 & 0.4773 & 0.4204 & 0.6948 & 0.3117 & 0.0552 & 0.1490 \\ 
\cmidrule{1-14}
\multirow{4}{*}{\shortstack[l]{Large-size\\Few-Shot \& CoT}} 
 & GPT-4o & 0.5227 & 0.2146 & 0.3182 & 0.2262 & 0.6688 & \underline{0.6902} & 0.4578 & 0.4564 & 0.7175 & 0.2727 & 0.0422 & 0.1500 \\
 & DeepSeek-V3.1 & 0.5909 & \textbf{0.3154} & 0.3149 & 0.2466 & 0.5974 & 0.6035 & 0.4578 & 0.4342 & 0.6753 & 0.2857 & 0.0617 & 0.1503 \\
 & Claude-haiku-4.5 & 0.5292 & 0.2676 & 0.2866 & 0.2120 & 0.5519 & 0.5404 & 0.5212 & 0.4063 & 0.6753 & 0.2662 & 0.0357 & 0.1550 \\
 & Qwen-Max & 0.4805 & 0.2212 & 0.2500 & 0.1887 & 0.6299 & 0.6398 & 0.4805 & 0.3938 & 0.6396 & 0.2630 & 0.0260 & 0.1596 \\
\midrule
\rowcolor{header_blue} \multicolumn{14}{c}{\textbf{Open-Source}} \\
\multirow{5}{*}{After SFT} 
 & Llama3.1-8b & 0.5812 & 0.2504 & 0.4026 & 0.2060 & 0.6429 & 0.6568 & 0.5325 & 0.3335 & 0.7273 & 0.4026 & 0.0974 & 0.1362 \\
 & Qwen3-8B & 0.5877 & 0.2607 & 0.4481 & 0.2536 & 0.6656 & 0.6507 & 0.5714 & 0.4247 & 0.7532 & 0.4221 & \textbf{0.1396} & 0.1275 \\
 & Qwen2.5-7B & 0.5390 & 0.2025 & \textbf{0.4870} & 0.2087 & 0.6136 & 0.5925 & 0.5455 & 0.3567 & 0.7403 & 0.3929 & 0.0974 & 0.1332 \\
 & DeepSeek-R1-Qwen3 & 0.5547 & 0.2352 & 0.2960 & 0.1795 & 0.6385 & 0.4728 & 0.4685 & 0.3975 & 0.6282 & 0.2852 & 0.0650 & 0.1356 \\
 & Mistral-7b & 0.3571 & 0.1519 & 0.1883 & 0.0991 & 0.4870 & 0.4950 & 0.2435 & 0.1728 & 0.0779 & 0.4058 & 0.0097 & 0.2002 \\
\midrule
\rowcolor{header_blue} \multicolumn{14}{c}{\textbf{Ours}} \\
\multirow{2}{*}{After SFT} 
 & HyCoLLM-Qwen3-8b & \underline{0.6169} & 0.2877 & \underline{0.4805} & \textbf{0.2826} & \textbf{0.7370} & \textbf{0.7449} & \underline{0.5844} & \underline{0.4848} & \textbf{0.8149} & \textbf{0.5065} & 0.1234 & \textbf{0.1165} \\
 & HyCoLLM-Llama3.1-8b & \textbf{0.6526} & \underline{0.2918} & 0.4643 & \underline{0.2677} & \underline{0.6753} & 0.6650 & \textbf{0.6201} & \textbf{0.4933} & \underline{0.8117} & \underline{0.5032} & \underline{0.1364} & \underline{0.1169} \\ 
\bottomrule
\end{tabular}
}
\caption{Performance comparison on the \textbf{CUT Dataset}. Rows are grouped by model type. The best results among all models are highlighted in \textbf{bold}, and the second-best results are \underline{underlined}. Note that for \textit{Hamming Loss}, lower is better ($\downarrow$).}
\label{tab:main_results}
\end{table*}

\begin{table*}[htpb]
\centering

\resizebox{\textwidth}{!}{
\begin{tabular}{ll | cc cc cc cc | ccc | c }
\toprule
\multicolumn{2}{c|}{\multirow{2}{*}{Model}} & 
\multicolumn{2}{c}{Emotion} & 
\multicolumn{2}{c}{Thinking} & 
\multicolumn{2}{c}{Stance} & 
\multicolumn{2}{c|}{Intent} & 
\multirow{2}{*}{PMA@2} & 
\multirow{2}{*}{PMA@3} & 
\multirow{2}{*}{PMA@4} & 
\multirow{2}{*}{\textit{Hamming Loss} $\downarrow$} \\
\multicolumn{2}{c|}{} & ACC & F1 & ACC & F1 & ACC & F1 & ACC & F1 & & & & \\ 
\midrule

\rowcolor{header_blue} \multicolumn{14}{c}{\textbf{Human}} \\
\multicolumn{2}{c|}{Averaged from 3 experts} & 0.7099 & 0.5095 & 0.6627 & 0.4770 & 0.8040 & 0.6296 & 0.7161 & 0.5357 & 0.9121 & 0.6890 & 0.3073 & 0.0819 \\
\midrule
\rowcolor{header_blue} \multicolumn{14}{c}{\textbf{Close-Source}} \\
\multirow{4}{*}{\shortstack[l]{Small-size\\Few-Shot \& CoT}} 
 & GPT4o-mini & 0.4449 & 0.3160 & 0.3392 & 0.2467 & 0.6498 & 0.6300 & 0.5220 & 0.4162 & 0.6872 & 0.2885 & 0.0507 & 0.1514 \\
 & Gemini2.5-Flash-Lite & 0.5297 & 0.3857 & 0.2967 & 0.2145 & 0.6418 & 0.6090 & 0.5846 & 0.4697 & 0.7253 & 0.3385 & 0.0418 & 0.1442 \\
 & Grok-4-fast & 0.3538 & 0.3350 & 0.3077 & 0.2702 & 0.6352 & 0.5513 & 0.5582 & 0.4907 & 0.6725 & 0.2330 & 0.0198 & 0.1589 \\
 & Qwen-Turbo & 0.3407 & 0.3229 & 0.3099 & 0.2009 & 0.5077 & 0.3535 & 0.5385 & 0.3281 & 0.5736 & 0.1956 & 0.0132 & 0.1687 \\ 
\cmidrule{1-14}
\multirow{4}{*}{\shortstack[l]{Large-size\\Few-Shot \& CoT}} 
 & GPT-4o & 0.5033 & \textbf{0.4679} & 0.3209 & 0.2349 & \textbf{0.6901} & \textbf{0.6826} & 0.4352 & 0.4116 & 0.6769 & 0.2923 & 0.0505 & 0.1519 \\
 & DeepSeek-V3.1 & 0.4840 & 0.4272 & 0.3158 & 0.2632 & 0.6476 & 0.6086 & 0.5675 & 0.4416 & 0.7032 & 0.3356 & 0.0525 & 0.1468 \\
 & Claude-haiku-4.5 & 0.3473 & 0.3867 & 0.3201 & 0.2744 & 0.6402 & 0.5774 & 0.5673 & 0.4100 & 0.6725 & 0.2549 & 0.0286 & 0.1569 \\
 & Qwen-Max & 0.4176 & 0.3693 & 0.2615 & 0.1797 & 0.5912 & 0.4319 & 0.5868 & 0.4171 & 0.6593 & 0.2505 & 0.0286 & 0.1566 \\
\midrule
\rowcolor{header_blue} \multicolumn{14}{c}{\textbf{Open-Source}} \\
\multirow{5}{*}{After SFT} 
 & Llama3.1-8b & 0.5231 & 0.3297 & 0.4286 & 0.1923 & 0.6220 & 0.5972 & 0.6110 & 0.4803 & 0.7341 & 0.3978 & 0.1099 & 0.1342 \\
 & Qwen3-8B & \underline{0.5714} & 0.3901 & 0.4505 & \underline{0.3851} & 0.6176 & 0.5773 & \underline{0.6615} & \textbf{0.5890} & \underline{0.8088} & 0.4154 & 0.1099 & \underline{0.1258} \\
 & Qwen2.5-7B & 0.5429 & 0.3853 & 0.4879 & 0.2528 & 0.6242 & 0.5944 & 0.6352 & 0.4883 & 0.7714 & 0.4396 & \underline{0.1165} & 0.1263 \\
 & DeepSeek-R1-Qwen3 & 0.4923 & 0.3607 & 0.3341 & 0.2227 & 0.5648 & 0.5487 & 0.5670 & 0.4848 & 0.6615 & 0.3165 & 0.0593 & 0.1509 \\
 & Mistral-7b & 0.3692 & 0.1025 & 0.1978 & 0.0632 & 0.2066 & 0.2104 & 0.1736 & 0.0865 & 0.2418 & 0.0352 & 0.0100 & 0.2025 \\
\midrule
\rowcolor{header_blue} \multicolumn{14}{c}{\textbf{Ours}} \\
\multirow{2}{*}{After SFT} 
 & HyCoLLM-Qwen3-8b & \textbf{0.6264} & \underline{0.4587} & \textbf{0.5341} & \textbf{0.3970} & \underline{0.6527} & \underline{0.6456} & \textbf{0.6637} & \underline{0.5813} & \textbf{0.8308} & \textbf{0.5209} & \textbf{0.1495} & \textbf{0.1128} \\
 & HyCoLLM-Llama3.1-8b & 0.5604 & 0.4028 & \underline{0.5033} & 0.2962 & 0.5516 & 0.4115 & 0.6505 & 0.5667 & 0.7560 & \underline{0.4462} & 0.1011 & 0.1278 \\ 
\bottomrule
\end{tabular}
}
\caption{Performance comparison on the \textbf{UE Dataset}. The best results are highlighted in \textbf{bold}, and the second-best are \underline{underlined}. For \textit{Hamming Loss}, lower is better ($\downarrow$).}
\label{tab:ue_results}
\end{table*}

\begin{table*}[htpb]
\centering

\resizebox{\textwidth}{!}{
\begin{tabular}{ll | cc cc cc cc | ccc | c }
\toprule
\multicolumn{2}{c|}{\multirow{2}{*}{Model}} & 
\multicolumn{2}{c}{Emotion} & 
\multicolumn{2}{c}{Thinking} & 
\multicolumn{2}{c}{Stance} & 
\multicolumn{2}{c|}{Intent} & 
\multirow{2}{*}{PMA@2} & 
\multirow{2}{*}{PMA@3} & 
\multirow{2}{*}{PMA@4} & 
\multirow{2}{*}{\textit{Hamming Loss} $\downarrow$} \\
\multicolumn{2}{c|}{} & ACC & F1 & ACC & F1 & ACC & F1 & ACC & F1 & & & & \\ 
\midrule

\rowcolor{header_blue} \multicolumn{14}{c}{\textbf{Human}} \\
\multicolumn{2}{c|}{Averaged from 3 experts} & 0.7036 & 0.5182 & 0.6050 & 0.4494 & 0.7789 & 0.7211 & 0.7004 & 0.5874 & 0.8932 & 0.6447 & 0.2627 & 0.0896 \\
\midrule
\rowcolor{header_blue} \multicolumn{14}{c}{\textbf{Close-Source}} \\
\multirow{4}{*}{\shortstack[l]{Small-size\\Few-Shot \& CoT}} 
 & GPT4o-mini & 0.4531 & 0.3803 & 0.2865 & 0.2214 & 0.3698 & 0.3814 & 0.5625 & 0.4120 & 0.5781 & 0.2083 & 0.0365 & 0.2009 \\
 & Gemini2.5-Flash-Lite & 0.4922 & 0.2251 & 0.3886 & 0.3366 & 0.3990 & 0.4002 & 0.6062 & 0.3621 & 0.5959 & 0.2694 & 0.0881 & 0.1825 \\
 & Grok-4-fast & 0.4249 & 0.2772 & 0.3212 & 0.2817 & 0.4093 & 0.4096 & 0.5907 & \textbf{0.5801} & 0.5855 & 0.2487 & 0.0155 & 0.1946 \\
 & Qwen-Turbo & 0.4508 & 0.2706 & 0.3886 & 0.2785 & 0.3212 & 0.2767 & 0.6425 & 0.4521 & 0.5751 & 0.2383 & 0.0829 & 0.1906 \\ 
\cmidrule{1-14}
\multirow{4}{*}{\shortstack[l]{Large-size\\Few-Shot \& CoT}} 
 & GPT-4o & 0.5208 & 0.2124 & 0.3385 & 0.3178 & 0.6198 & 0.5563 & 0.5573 & 0.3965 & 0.6667 & 0.3333 & 0.0781 & 0.1698 \\
 & DeepSeek-V3.1 & 0.5130 & 0.2420 & 0.3281 & 0.2818 & 0.4688 & 0.4746 & 0.5729 & 0.4016 & 0.6373 & 0.2539 & 0.0622 & 0.1820 \\
 & Claude-haiku-4.5 & 0.4508 & 0.2928 & 0.3005 & 0.2296 & 0.3782 & 0.3960 & 0.5492 & 0.2668 & 0.5648 & 0.1865 & 0.0363 & 0.2005 \\
 & Qwen-Max & 0.4611 & 0.3376 & 0.2953 & 0.2485 & 0.5130 & 0.4899 & 0.6218 & 0.4445 & 0.6477 & 0.3212 & 0.0311 & 0.1825 \\
\midrule
\rowcolor{header_blue} \multicolumn{14}{c}{\textbf{Open-Source}} \\
\multirow{5}{*}{After SFT} 
 & Llama3.1-8b & 0.4767 & \textbf{0.3968} & 0.2798 & 0.1306 & 0.6528 & 0.4088 & 0.4663 & 0.1915 & 0.6528 & 0.2435 & 0.0311 & 0.1845 \\
 & Qwen3-8B & 0.6010 & 0.3175 & 0.3938 & 0.2911 & 0.6632 & 0.3717 & 0.6010 & 0.4848 & 0.8342 & 0.3627 & 0.0777 & 0.1514 \\
 & Qwen2.5-7B & 0.5429 & \underline{0.3853} & 0.4879 & 0.2528 & 0.6242 & \textbf{0.5944} & 0.6352 & 0.4883 & 0.7714 & 0.4396 & 0.1165 & 0.1263 \\
 & DeepSeek-R1-Qwen3 & 0.5389 & 0.2664 & 0.3834 & 0.2799 & 0.5233 & 0.4192 & 0.5907 & 0.4033 & 0.7098 & 0.3005 & 0.0674 & 0.1687 \\
 & Mistral-7b & 0.3316 & 0.1061 & 0.2073 & 0.0927 & 0.6269 & 0.2807 & 0.3420 & 0.0983 & 0.5026 & 0.1036 & 0.0207 & 0.2120 \\
\midrule
\rowcolor{header_blue} \multicolumn{14}{c}{\textbf{Ours}} \\
\multirow{2}{*}{After SFT} 
 & HyCoLLM-Qwen3-8b & \textbf{0.6425} & 0.3052 & \textbf{0.5855} & \underline{0.4805} & \underline{0.7202} & 0.5849 & \underline{0.6943} & \underline{0.5759} & \textbf{0.8860} & \textbf{0.5803} & \textbf{0.1917} & \textbf{0.1176} \\
 & HyCoLLM-Llama3.1-8b & \underline{0.6269} & 0.2920 & \underline{0.5596} & \textbf{0.5674} & \textbf{0.7254} & \underline{0.5863} & \textbf{0.7047} & 0.5577 & \textbf{0.8860} & \underline{0.5648} & \textbf{0.1917} & \underline{0.1198} \\ 
\bottomrule
\end{tabular}
}
\caption{Performance comparison on the \textbf{DEI Dataset}. The best results are highlighted in \textbf{bold}, and the second-best are \underline{underlined}. For \textit{Hamming Loss}, lower is better ($\downarrow$).}
\label{tab:dei_results}
\end{table*}

\subsection{Detailed generalization analysis}
\label{app: generalization}

The complete cross-dataset generalization results, where models trained on the CUT dataset are tested on the entirely unseen FRIR dataset, are detailed in Table \ref{tab:frir_generalization_full}. These comprehensive metrics strongly reinforce the core findings in the main text: HyCoLLM, which embeds cognitive geometric priors, significantly enhances the model's generalization ability. Specifically, HyCoLLM-Llama3.1-8b achieves the lowest \textit{Hamming Loss} ($\downarrow 0.1248$) among all non-human models and performs excellently in all composite metrics (PMA@2, PMA@3, PMA@4). In particular, the leading position in the PMA@4 accuracy (0.1117), which measures the model's ability to correctly predict all four dimensions simultaneously, strongly demonstrates that the Hyperbolic Cognitive Network (HCN) successfully decouples complex, multi-dimensional cognitive states using the Poincaré sphere. This provides large language models with a more essential, topic-independent cognitive understanding, thereby effectively resisting overfitting in cross-domain transfer.

In terms of dimensional performance, HyCoLLM stands out in core mental state dimensions such as Emotion, Thinking, and Intent. For example, HyCoLLM-Llama3.1-8b leads in Thinking ACC (0.5978) and Intent F1 (0.6147), indicating that hyperbolic space effectively captures the hierarchical relationships between these mental states that are less dependent on topic semantics. However, as discussed in the main text, the Stance dimension is an exception, where HyCoLLM's performance is relatively mediocre. This is an important and expected finding: unlike intrinsic mental states such as emotion and thinking, the stance labels in the FRIR dataset (specific claims regarding opposing topics) are semantically fundamentally different from those in the CUT training set, and their correct prediction is highly dependent on topic-specific semantic information. This divergence in the definition of target concepts highlights the limitations of cognitive geometric priors, i.e., when the target dimension is primarily determined by superficial, domain-specific semantics, the transferability of its structured understanding is restricted. Despite this shortcoming in the Stance dimension, HyCoLLM still outperforms all SFT baselines and most powerful closed-source LLMs (such as GPT-4o and DeepSeek-V3.1) in overall multi-dimensional alignment metrics (composite metrics and \textit{Hamming Loss}), confirming its structural advantages in encoding human-like cognitive structures.

\begin{table*}[tb]
\centering

\resizebox{\textwidth}{!}{%
\begin{tabular}{ll | cc cc cc cc | ccc | c }
\toprule
\multicolumn{2}{c|}{\multirow{2}{*}{Model}} & 
\multicolumn{2}{c}{Emotion} & 
\multicolumn{2}{c}{Thinking} & 
\multicolumn{2}{c}{Stance} & 
\multicolumn{2}{c|}{Intent} & 
\multirow{2}{*}{PMA@2} & 
\multirow{2}{*}{PMA@3} & 
\multirow{2}{*}{PMA@4} & 
\multirow{2}{*}{\textit{Hamming Loss} $\downarrow$} \\
\multicolumn{2}{c|}{} & ACC & F1 & ACC & F1 & ACC & F1 & ACC & F1 & & & & \\ 
\midrule

\rowcolor[HTML]{EAF4F7} \multicolumn{14}{c}{\textbf{Human}} \\
\multicolumn{2}{c|}{Averaged from 3 experts} & 0.6663 & 0.5732 & 0.7236 & 0.5113 & 0.7871 & 0.3971 & 0.7882 & 0.6456 & 0.9266 & 0.7178 & 0.3331 & 0.0688 \\
\midrule
\rowcolor[HTML]{EAF4F7} \multicolumn{14}{c}{\textbf{Close-Source}} \\
\multirow{4}{*}{\shortstack[l]{Small-size\\Few-Shot \& CoT}} 
 & GPT4o-mini & 0.5559 & 0.4182 & 0.2514 & 0.1846 & 0.4637 & 0.4382 & 0.4888 & 0.3853 & 0.6089 & 0.2123 & 0.0335 & 0.1659 \\
 & Gemini2.5-Flash-Lite & 0.5574 & 0.3663 & 0.3389 & 0.2513 & \textbf{0.5070} & \textbf{0.4973} & 0.6275 & 0.4768 & 0.6891 & 0.3389 & 0.0616 & 0.1458 \\
 & Grok-4-fast & 0.5475 & \textbf{0.5089} & 0.3101 & 0.2718 & 0.4804 & \underline{0.4763} & 0.5112 & 0.4659 & 0.6229 & 0.2486 & 0.0559 & 0.1593 \\
 & Qwen-Turbo & 0.4665 & 0.2721 & 0.3575 & 0.2217 & 0.4777 & 0.3208 & 0.4441 & 0.3422 & 0.6006 & 0.2430 & 0.0559 & 0.1653 \\ 
\cmidrule{1-14}
\multirow{4}{*}{\shortstack[l]{Large-size\\Few-Shot \& CoT}} 
 & GPT-4o & \underline{0.5866} & 0.4445 & 0.2654 & 0.2097 & 0.4665 & 0.4274 & 0.4218 & 0.3789 & 0.6006 & 0.2123 & 0.0363 & 0.1674 \\
 & DeepSeek-V3.1 & 0.5359 & 0.4271 & 0.3725 & 0.3057 & \underline{0.4874} & 0.4689 & 0.4678 & 0.4341 & 0.6779 & 0.3137 & 0.0532 & 0.1508 \\
 & claude-haiku-4.5 & 0.4510 & 0.3745 & 0.3417 & 0.2848 & 0.4650 & 0.4657 & 0.5686 & 0.4057 & 0.6162 & 0.2717 & 0.0504 & 0.1610 \\
 & Qwen-Max & 0.5279 & 0.3383 & 0.2598 & 0.2046 & 0.4832 & 0.3412 & 0.5140 & 0.3780 & 0.6006 & 0.2430 & 0.0447 & 0.1626 \\
\midrule
\rowcolor[HTML]{EAF4F7} \multicolumn{14}{c}{\textbf{Open-Source (SFT Baseline)}} \\
\multirow{5}{*}{After SFT} 
 & Llama3.1-8b & 0.4944 & 0.2067 & 0.3799 & 0.2040 & 0.3911 & 0.2734 & 0.5279 & 0.3280 & 0.6117 & 0.2486 & 0.0531 & 0.1592 \\
 & Qwen3-8B & 0.5084 & 0.3363 & \underline{0.5782} & \textbf{0.3484} & 0.4218 & 0.3305 & 0.6229 & 0.3911 & 0.7263 & \underline{0.3827} & 0.0922 & 0.1384 \\
 & Qwen2.5-7B & 0.4637 & 0.2604 & 0.4749 & 0.2141 & 0.4162 & 0.3279 & 0.5140 & 0.3229 & 0.6257 & 0.2849 & 0.0782 & 0.1575 \\
 & DeepSeek-R1-Qwen3 & 0.5439 & 0.3774 & 0.2885 & 0.2049 & 0.4638 & 0.3306 & 0.4137 & 0.3433 & 0.4811 & 0.1950 & 0.0409 & 0.1401 \\
 & Mistral-7b & 0.3240  & 0.1061  & 0.1285  & 0.0521  & 0.3631  & 0.2812  & 0.2318  & 0.1298  & 0.2961  & 0.0223  & 0.0028  & 0.2187  \\
\midrule
\rowcolor[HTML]{EAF4F7} \multicolumn{14}{c}{\textbf{Ours (HyCoLLM)}} \\
\multirow{2}{*}{After SFT} 
 & HyCoLLM-Qwen3-8b & 0.5838 & 0.3726 & 0.5056 & 0.3285 & 0.4190 & 0.3162 & \underline{0.6704} & \underline{0.5042} & \underline{0.7458} & \underline{0.3827} & \underline{0.1061} & \underline{0.1349} \\
 & HyCoLLM-Llama3.1-8b & \textbf{0.5978} & \underline{0.4546} & \textbf{0.5978} & \underline{0.3371} & 0.3743 & 0.1867 & \textbf{0.6983} & \textbf{0.6147} & \textbf{0.7737} & \textbf{0.4274} & \textbf{0.1117} & \textbf{0.1248} \\ 
\bottomrule
\end{tabular}%
}
\caption{Full cross-dataset generalization performance, trained on CUT and tested on the unseen FRIR dataset. The best results among non-human models are highlighted in \textbf{bold}, and the second-best are \underline{underlined}. For \textit{Hamming Loss}, lower is better ($\downarrow$).}
\label{tab:frir_generalization_full}
\end{table*}

\

\subsection{Detailed Ablation Study on Sub-datasets}
\label{app:ablation_sub}

\begin{table*}[htpb]
\centering

\resizebox{\textwidth}{!}{
\begin{tabular}{l | cc cc cc cc | ccc | c }
\toprule
\multirow{2}{*}{Method} & \multicolumn{2}{c}{Emotion} & \multicolumn{2}{c}{Thinking} & \multicolumn{2}{c}{Stance} & \multicolumn{2}{c|}{Intent} & \multirow{2}{*}{PMA@2} & \multirow{2}{*}{PMA@3} & \multirow{2}{*}{PMA@4} & \multirow{2}{*}{H-Loss $\downarrow$} \\
 & ACC & F1 & ACC & F1 & ACC & F1 & ACC & F1 & & & & \\ 
\midrule
\textbf{HyCoLLM (Full)} & \textbf{0.6169} & \textbf{0.2877} & \textbf{0.4805} & \textbf{0.2826} & \textbf{0.7370} & \textbf{0.7449} & 0.5844 & 0.4848 & \textbf{0.8149} & \textbf{0.5065} & \textbf{0.1234} & \textbf{0.1165} \\
\rowcolor{header_blue} \multicolumn{13}{c}{\textbf{Phase I Ablations (HCN)}} \\
w/o Hyperbolic & 0.6136 & 0.2223 & 0.4188 & 0.2633 & 0.7305 & 0.7338 & \textbf{0.5974} & 0.4753 & 0.7955 & 0.4643 & \textbf{0.1234} & 0.1212 \\
w/o Geo-Regularization & \textbf{0.6169} & 0.2455 & 0.4416 & 0.2551 & 0.7078 & 0.7110 & \textbf{0.5974} & \textbf{0.5028} & 0.7987 & 0.4708 & 0.1266 & 0.1209 \\
w/o Cross-Interaction & 0.6039 & 0.2257 & 0.4416 & 0.2811 & 0.7045 & 0.7058 & 0.5942 & 0.4924 & 0.7922 & 0.4578 & 0.1169 & 0.1224 \\
\rowcolor{header_blue} \multicolumn{13}{c}{\textbf{Phase II Ablations (HGAT)}} \\
w/o Soft Prompt & 0.6104 & 0.2281 & 0.4318 & 0.2705 & 0.7045 & 0.7080 & 0.5942 & 0.4841 & 0.7955 & 0.4610 & 0.1136 & 0.1227 \\
w/o SCT Loss & 0.6364 & 0.2472 & 0.4643 & \textbf{0.2879} & 0.7143 & 0.7252 & 0.5877 & 0.4940 & 0.7987 & 0.4708 & 0.1266 & 0.1209 \\
w/o Cognitive Anchor & 0.6039 & 0.2208 & 0.4351 & 0.2829 & 0.7078 & 0.7089 & 0.5942 & 0.4793 & 0.8052 & 0.4481 & 0.1201 & 0.1227 \\
\bottomrule
\end{tabular}
}
\caption{Ablation study on the \textbf{CogBiasESC} dataset. \textbf{Bold} indicates the best performance (Full Model).}
\label{tab:ablation_cut}
\end{table*}

\begin{table*}[htpb]
\centering

\resizebox{\textwidth}{!}{
\begin{tabular}{l | cc cc cc cc | ccc | c }
\toprule
\multirow{2}{*}{Method} & \multicolumn{2}{c}{Emotion} & \multicolumn{2}{c}{Thinking} & \multicolumn{2}{c}{Stance} & \multicolumn{2}{c|}{Intent} & \multirow{2}{*}{PMA@2} & \multirow{2}{*}{PMA@3} & \multirow{2}{*}{PMA@4} & \multirow{2}{*}{H-Loss $\downarrow$} \\
 & ACC & F1 & ACC & F1 & ACC & F1 & ACC & F1 & & & & \\ 
\midrule
\textbf{HyCoLLM (Full)} & \textbf{0.6264} & \textbf{0.4587} & \textbf{0.5341} & \textbf{0.3970} & \textbf{0.6527} & \textbf{0.6456} & \textbf{0.6637} & \textbf{0.5813} & \textbf{0.8308} & \textbf{0.5209} & \textbf{0.1495} & \textbf{0.1128} \\
\rowcolor{header_blue} \multicolumn{13}{c}{\textbf{Phase I Ablations (HCN)}} \\
w/o Hyperbolic & 0.5956 & 0.3567 & 0.4945 & 0.3488 & 0.6374 & 0.6165 & 0.6571 & 0.5284 & 0.8110 & 0.4923 & 0.1275 & 0.1197 \\
w/o Geo-Regularization & 0.5890 & 0.3877 & 0.4835 & 0.3682 & 0.6264 & 0.6053 & 0.6615 & 0.5401 & 0.8044 & 0.4615 & 0.1363 & 0.1214 \\
w/o Cross-Interaction & 0.5956 & 0.3976 & 0.5121 & 0.3843 & 0.6352 & 0.6184 & 0.6440 & 0.5171 & 0.7956 & 0.4945 & 0.1341 & 0.1195 \\
\rowcolor{header_blue} \multicolumn{13}{c}{\textbf{Phase II Ablations (HGAT)}} \\
w/o Soft Prompt & 0.6066 & 0.3839 & 0.4835 & 0.3365 & 0.6110 & 0.5944 & \textbf{0.6637} & 0.5276 & 0.7956 & 0.4703 & 0.1407 & 0.1211 \\
w/o SCT Loss & 0.6154 & 0.4437 & 0.4923 & 0.3834 & 0.6264 & 0.6124 & 0.6374 & 0.5150 & 0.8066 & 0.4703 & 0.1297 & 0.1206 \\
w/o Cognitive Anchor & 0.5956 & 0.3840 & 0.5011 & 0.3023 & 0.6176 & 0.6034 & 0.6593 & 0.5175 & 0.7956 & 0.4703 & 0.1407 & 0.1205 \\
\bottomrule
\end{tabular}
}
\caption{Ablation study on the \textbf{UE} dataset.}
\label{tab:ablation_ue}
\end{table*}

\begin{table*}[htpb]
\centering

\resizebox{\textwidth}{!}{
\begin{tabular}{l | cc cc cc cc | ccc | c }
\toprule
\multirow{2}{*}{Method} & \multicolumn{2}{c}{Emotion} & \multicolumn{2}{c}{Thinking} & \multicolumn{2}{c}{Stance} & \multicolumn{2}{c|}{Intent} & \multirow{2}{*}{PMA@2} & \multirow{2}{*}{PMA@3} & \multirow{2}{*}{PMA@4} & \multirow{2}{*}{H-Loss $\downarrow$} \\
 & ACC & F1 & ACC & F1 & ACC & F1 & ACC & F1 & & & & \\ 
\midrule
\textbf{HyCoLLM (Full)} & 0.6425 & 0.3052 & \textbf{0.5855} & \textbf{0.4805} & \textbf{0.7202} & \textbf{0.5849} & \textbf{0.6943} & \textbf{0.5759} & \textbf{0.8860} & \textbf{0.5803} & \textbf{0.1917} & \textbf{0.1176} \\
\rowcolor{header_blue} \multicolumn{13}{c}{\textbf{Phase I Ablations (HCN)}} \\
w/o Hyperbolic & \textbf{0.6632} & 0.3736 & 0.4473 & 0.3616 & 0.6632 & 0.4473 & 0.6425 & 0.4244 & 0.8497 & 0.4922 & 0.1140 & 0.1361 \\
w/o Geo-Regularization & 0.6477 & \textbf{0.4216} & 0.4301 & 0.3548 & 0.6736 & 0.4421 & 0.6321 & 0.4510 & 0.8187 & 0.4819 & 0.0984 & 0.1403 \\
w/o Cross-Interaction & 0.6528 & 0.2461 & 0.4560 & 0.3596 & 0.6736 & 0.4487 & 0.6477 & 0.4335 & 0.8394 & 0.5130 & 0.0984 & 0.1361 \\
\rowcolor{header_blue} \multicolumn{13}{c}{\textbf{Phase II Ablations (HGAT)}} \\
w/o Soft Prompt & 0.6321 & 0.2384 & 0.4456 & 0.3808 & 0.6528 & 0.4375 & 0.6425 & 0.4228 & 0.8083 & 0.4819 & 0.0984 & 0.1406 \\
w/o SCT Loss & 0.6166 & 0.3680 & 0.4819 & 0.3866 & 0.7150 & 0.5299 & 0.6477 & 0.4363 & 0.8394 & 0.5233 & 0.1192 & 0.1331 \\
w/o Cognitive Anchor & 0.6528 & 0.2528 & 0.4560 & 0.3608 & 0.6684 & 0.4508 & 0.6580 & 0.4377 & 0.8290 & 0.5181 & 0.1036 & 0.1352 \\
\bottomrule
\end{tabular}
}
\caption{Ablation study on the \textbf{DEI} dataset.}
\label{tab:ablation_dei}
\end{table*}

In this section, we provide a granular analysis of the ablation studies for the CUT, UE, and DEI datasets. The results (Tables \ref{tab:ablation_cut}, \ref{tab:ablation_ue}, and \ref{tab:ablation_dei}) reveal that the contribution of each component varies significantly depending on the cognitive complexity of the domain.

\subsubsection{Ablation on CUT Dataset (Table \ref{tab:ablation_cut})}
The CUT dataset represents a relatively rational and structured domain. Consequently, the impact of geometric priors is more subtle but still critical for consistency.
\paragraph{Phase I: Geometry vs. Structure.} 
Interestingly, replacing the Poincaré ball with Euclidean space (\textit{w/o Hyperbolic}) results in a minimal drop in \textit{PMA@4} accuracy (maintained at 0.1234) and even a slight increase in Intent Accuracy. This suggests that for economic debates with clearer logical boundaries, Euclidean space is partially sufficient. However, the \textit{Thinking F1} score drops notably (0.2826 $\to$ 0.2633), indicating that even in rational discourse, the hierarchical nature of cognitive processes (System 1 vs. System 2) is better preserved in hyperbolic space. The removal of \textit{Cross-Interaction} leads to the lowest Stance F1 (0.7058), confirming that economic stance is heavily dependent on the interplay between thinking style and emotion.

\paragraph{Phase II: The Necessity of Alignment.}
The \textit{Soft Prompt} proves indispensable here. Its removal causes the sharpest decline in \textit{PMA@4} accuracy (to 0.1136) among all variants. This implies that even if the geometric prior is well-constructed, the LLM requires explicit "soft guidance" to map these structured features into its generation space. Without the \textit{SCT Loss}, the model fails to maintain the topological consistency of the Intent dimension (F1 drops from 0.4848 to 0.4841), proving that alignment tuning is crucial for precise intent recognition.

\subsubsection{Ablation on UE Dataset (Table \ref{tab:ablation_ue})}
In the polarized political domain, the "Cognitive Crowding" is more severe, and the benefits of our framework become more pronounced.
\paragraph{Phase I: Hierarchy Matters.} 
Unlike CUT, the \textit{w/o Hyperbolic} setting in UE leads to a clear degradation in \textit{PMA@4} accuracy (0.1495 $\to$ 0.1275) and a significant drop in Thinking F1 (0.3970 $\to$ 0.3488). This validates our hypothesis that political ideologies form a latent hierarchy (e.g., broad liberal values branching into specific policy stances). Euclidean space suffers from crowding when modeling these trees, causing the model to lose track of the nuanced thinking patterns underpinning political stances.

\paragraph{Phase II: Bridging the Gap.} 
The absence of \textit{Soft Prompts} causes a substantial performance drop (PMA@4 $\to$ 0.1407). Furthermore, removing the \textit{SCT Loss} leads to a higher \textit{Hamming Loss} (0.1128 $\to$ 0.1206), suggesting that without topological constraints, the LLM tends to generate semantically plausible but cognitively misaligned responses (e.g., predicting the correct Stance but the wrong underlying Emotion).

\subsubsection{Ablation on DEI Dataset (Table \ref{tab:ablation_dei})}
The DEI dataset, characterized by abstract values and high conflict, serves as the strongest proof of concept for HyCoLLM.
\paragraph{Phase I: The Criticality of Hyperbolic Space.} 
The results here are striking. Removing the hyperbolic geometry (\textit{w/o Hyperbolic}) causes a catastrophic collapse in holistic understanding: \textit{PMA@4} accuracy plummets by nearly 40\% (0.1917 $\to$ 0.1140), and Thinking F1 drops massively (0.4805 $\to$ 0.3616).
It is worth noting that while \textit{w/o Hyperbolic} achieves higher Emotion accuracy, it fails to connect this emotion to the correct thinking style and stance (hence the low PMA@4). This phenomenon perfectly illustrates the "crowding problem": Euclidean models may cluster surface-level sentiments well but fail to disentangle the complex, deep-seated value systems that drive DEI debates. Only hyperbolic space provides the exponential volume required to organize these entangled cognitive states.

\paragraph{Phase II: Alignment under Complexity.} 
In this complex domain, the \textit{Soft Prompt} is the lifeline of the model. Its removal reduces the \textit{PMA@4} accuracy to single digits (0.0984), comparable to the baseline models. This confirms that for abstract social concepts, the LLM cannot effectively infer the cognitive context from raw text alone; it heavily relies on the high-fidelity geometric signals injected via our soft prompts to navigate the "value maze" of DEI discussions.

\section{Training Dynamics Analysis}
\label{app:training_dynamics}

\begin{figure}[tb]
    \centering
    \includegraphics[width=\linewidth]{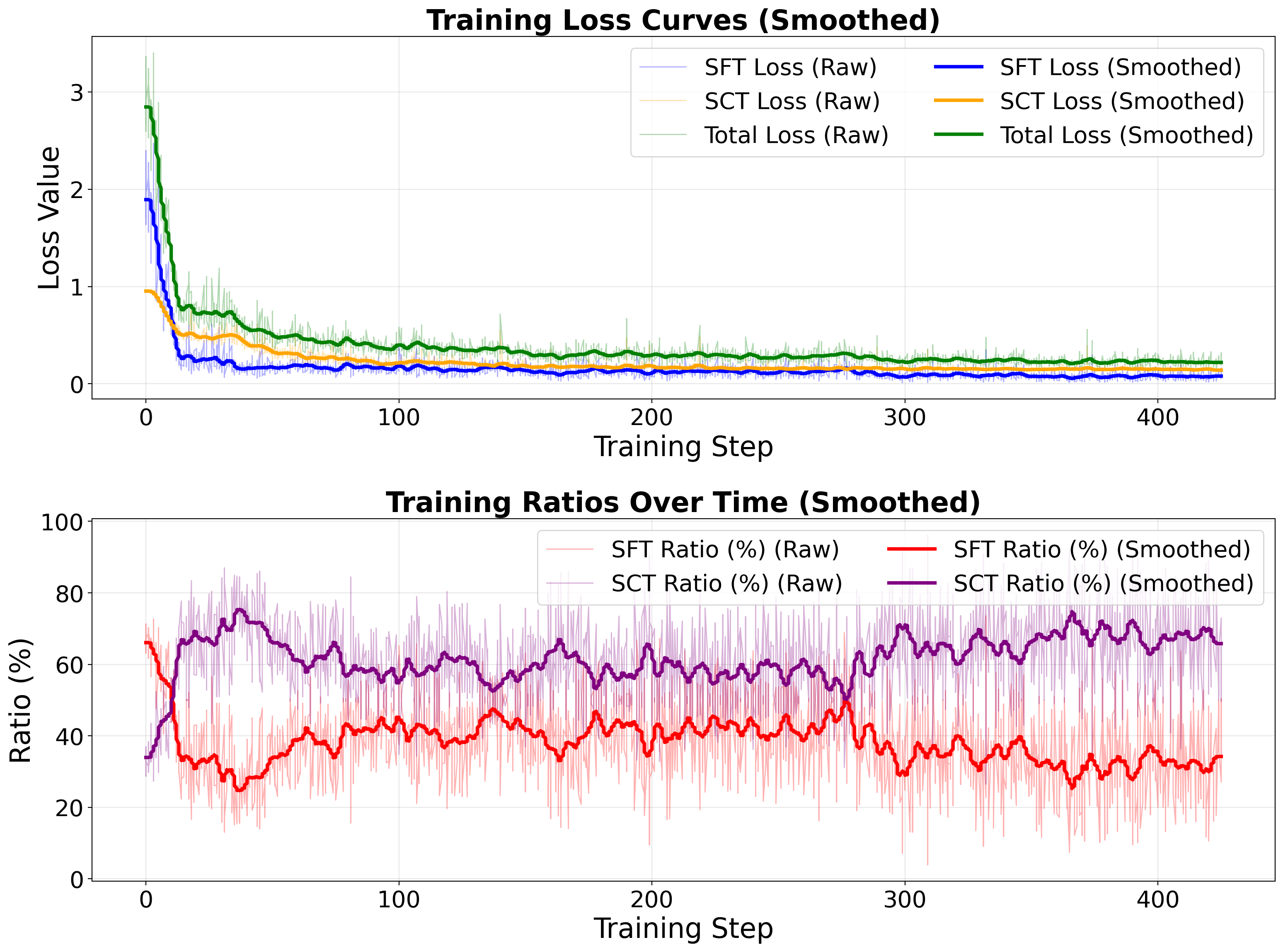} 
    \caption{Visualization of training dynamics. \textbf{Top:} The raw and smoothed loss curves for SFT (generation), SCT (alignment), and Total Loss. \textbf{Bottom:} The relative ratio (\%) of the SFT and SCT loss components over time. The curves demonstrate stable convergence and persistent geometric guidance.}
    \label{fig:training_loss}
\end{figure}

To further validate the stability of our Hyperbolic-Guided Alignment Tuning (HGAT), we visualize the training loss curves in Figure~\ref{fig:training_loss}. The data is recorded from the training process of HyCoLLM-Qwen3-8B with the optimal hyperparameters ($\lambda=1.0, L=4$).

\paragraph{Convergence Stability.}
As shown in the top panel of Figure~\ref{fig:training_loss}, the Total Loss (green line) exhibits a steep initial descent followed by a smooth convergence, indicating that the optimization landscape is stable despite the complexity of the hyperbolic constraints. Notably, both the SFT Loss (blue) and SCT Loss (orange) decrease synchronously. This synchronization confirms that the geometric alignment goal does not conflict with the semantic generation objective; rather, they are mutually compatible, allowing the model to learn cognitive structures without sacrificing language fluency.

\paragraph{Persistent Geometric Guidance.}
The bottom panel illustrates the evolution of the loss composition ratios. An interesting pattern emerges: while the SFT loss drops rapidly in the early stages (as the model quickly adapts to the output format), the SCT loss ratio (purple line) gradually rises and stabilizes around 60-70\%. This phenomenon suggests that as the model masters the basic textual patterns, the \textit{Semantic-Cognitive Topology} loss becomes the dominant gradient signal. This ensures that in the later stages of fine-tuning, the model focuses heavily on refining the subtle topological alignment of its feature space, preventing the "feature collapse" often seen in standard SFT.

\section{Human Evaluation Prompt}
\label{app:human_eval_prompt}

\begin{figure*}[tb]
    \centering
    \includegraphics[width=0.95\linewidth]{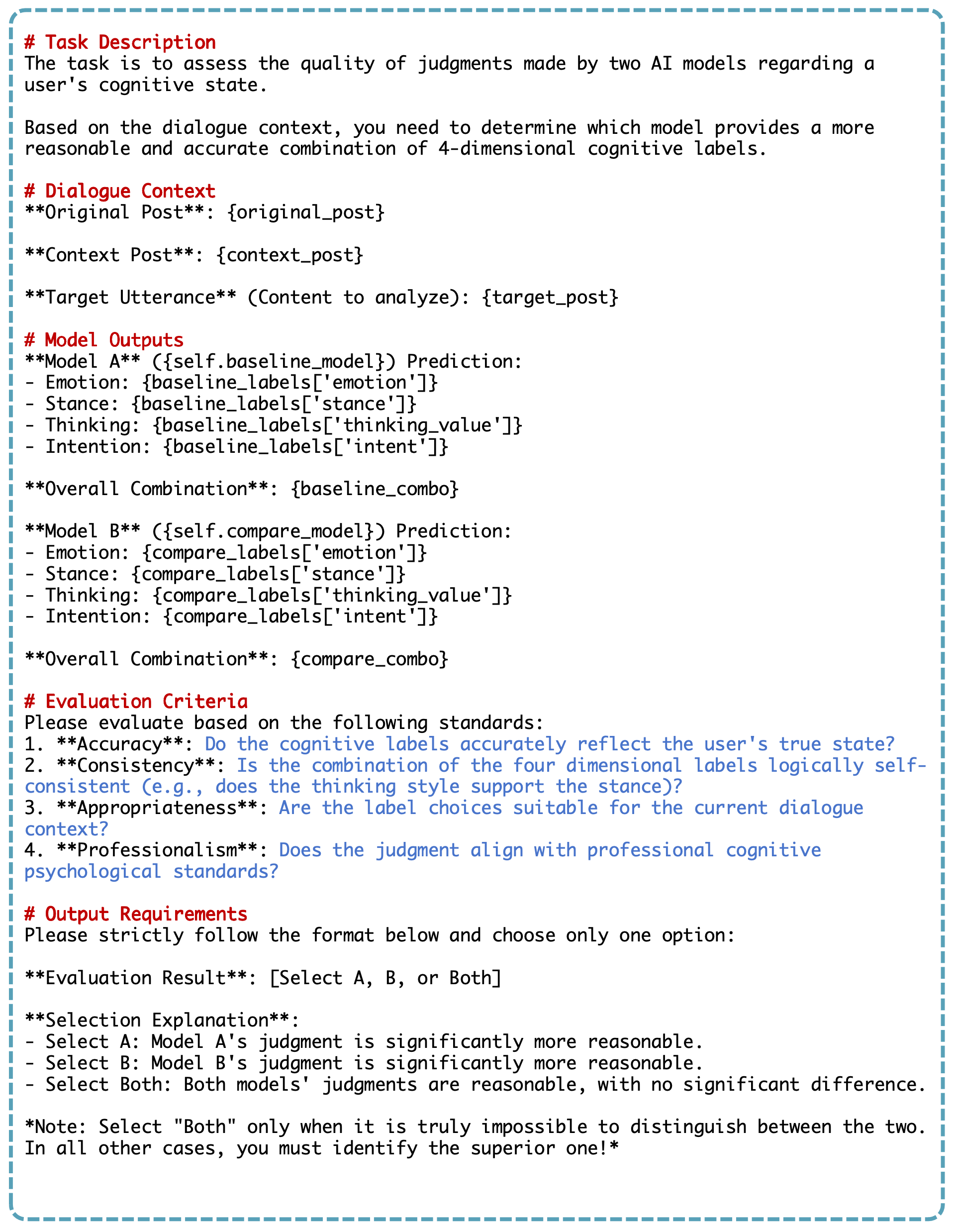}
    \caption{The instruction interface provided to human evaluators for pairwise comparison.}
    \label{fig:eval_prompt}
\end{figure*}

Figure~\ref{fig:eval_prompt} presents the detailed instructions and criteria used for the human evaluation. Experts were explicitly instructed to evaluate the "holistic combination" of the four cognitive dimensions rather than individual labels.

\section{Inference Prompt}

\begin{figure*}[tb]
    \centering
    \includegraphics[width=0.95\linewidth]{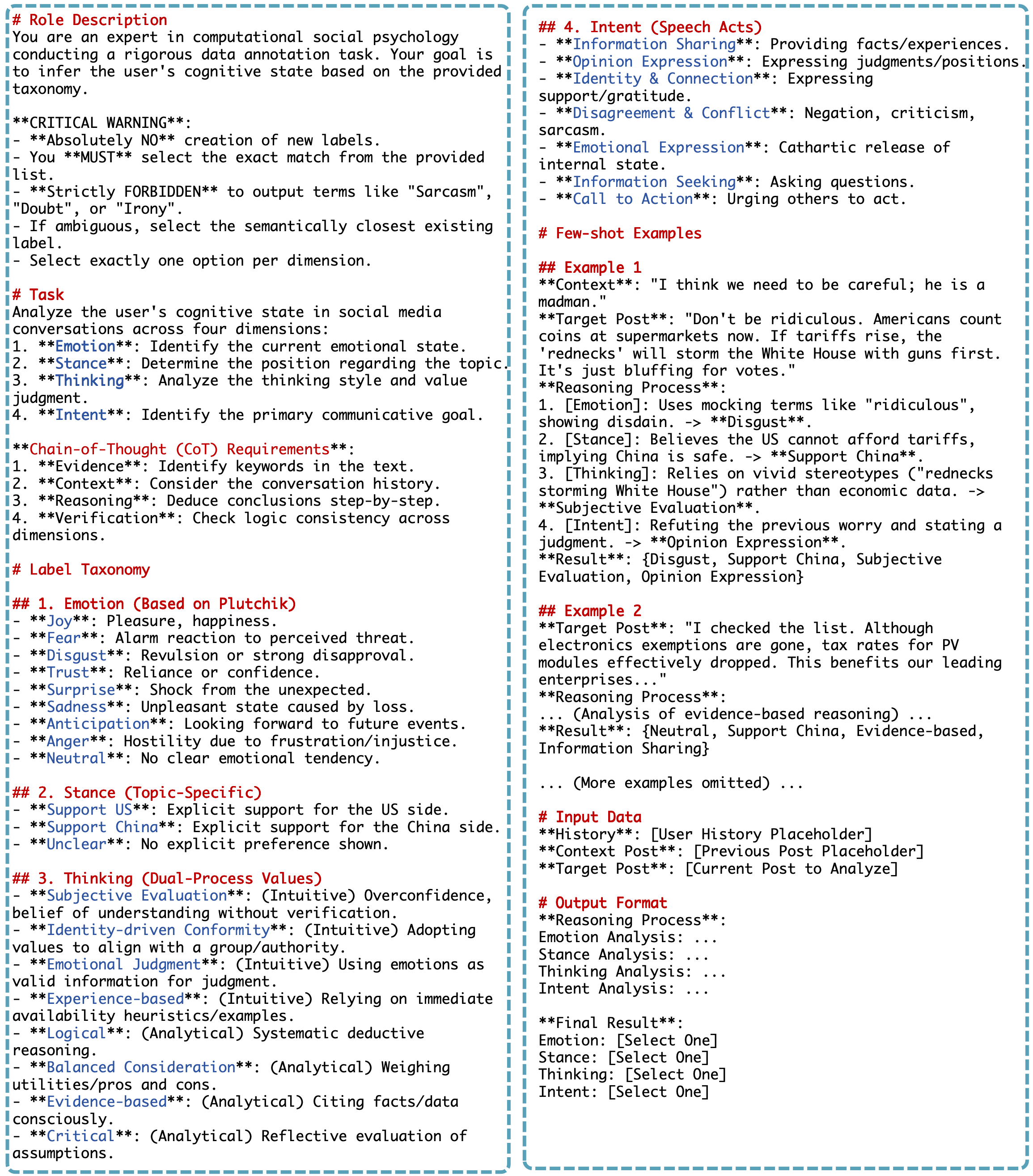}
    \caption{The prompt template used for closed-source LLMs (e.g., GPT-4o, Claude-4.5). The prompt employs a Chain-of-Thought (CoT) and Few-Shot strategy to guide the model in reasoning across four cognitive dimensions. Note that for our fine-tuned models (HyCoLLM and SFT baselines), the "Few-shot Examples" and detailed "CoT Requirements" sections were removed during inference to rely on the models' internalized parameter knowledge and to optimize inference latency.}
    \label{fig:inference_prompt}
\end{figure*}

\paragraph{Prompting Strategies.}
To ensure a fair comparison, we tailored the prompting strategies based on the model type:  Since we cannot access the weights of models like GPT-4o and Gemini, we employed a robust Few-shot + CoT prompt (illustrated in Figure~\ref{fig:inference_prompt}). This prompt includes detailed task definitions, strict output constraints, and three domain-specific examples with step-by-step reasoning paths to mitigate cognitive crowding and enforce label consistency. For HyCoLLM and other open-source baselines, we applied Supervised Fine-Tuning (SFT). During the inference phase for these models, we utilized a Zero-shot prompt structure. We stripped the few-shot examples and explicit reasoning instructions, retaining only the task description and label taxonomy. This approach validates the effectiveness of the fine-tuning process, demonstrating that HyCoLLM has successfully internalized the cognitive geometric prior and alignment constraints without needing lengthy in-context guidance. 
\end{document}